\pdfoutput=1

\PassOptionsToPackage{table,xcdraw}{xcolor}
\documentclass[11pt]{article}

\usepackage[preprint]{acl}

\usepackage{times}
\usepackage{latexsym}

\usepackage[T1]{fontenc}

\usepackage[utf8]{inputenc}

\usepackage{microtype}

\usepackage{inconsolata}

\usepackage{graphicx}
\usepackage{amsmath}
\usepackage{booktabs}
\usepackage{graphicx}
\usepackage{adjustbox}
\usepackage{lscape}
\usepackage{multirow}
\usepackage{adjustbox}
\usepackage{pifont} 
\usepackage{arydshln}
\usepackage{amssymb}
\usepackage{subcaption}

\usepackage{booktabs}      
\usepackage{siunitx}       
\usepackage{arydshln}      

\newcommand{\cmark}{\textcolor{green}{\ding{51}}} 
\newcommand{\xmark}{\textcolor{red}{\ding{55}}} 
\newcommand{\tcite}[1]{{\fontsize{5}{10}\selectfont\cite{#1}}}

\title{MAGIC-VQA: Multimodal And Grounded Inference \\ with Commonsense Knowledge for Visual Question Answering}

\author{
  Shuo Yang$^1$, Soyeon Caren Han$^{1}$\thanks{*Corresponding author. Caren.Han@unimelb.edu.au}, 
  Siwen Luo$^2$, Eduard Hovy$^1$ \\
  $^1$The University of Melbourne, $^2$The University of Western Australia\\
}

\date{}

\begin{document}
\maketitle
\begin{abstract}

Visual Question Answering (VQA) requires reasoning across visual and textual modalities, yet Large Vision-Language Models (LVLMs) often lack integrated commonsense knowledge, limiting their robustness in real-world scenarios. To address this, we introduce MAGIC-VQA, a novel framework that enhances VQA by systematically integrating  commonsense knowledge with LVLMs.
MAGIC-VQA employs a three-stage process: (1) Explicit Knowledge Integration from external sources, (2) By-Type Post-Processing for contextual refinement, and (3) Implicit Knowledge Augmentation using a Graph Neural Network (GNN) for structured reasoning. While GNNs bring greater depth to structured inference, they enable superior relational inference beyond LVLMs. MAGIC-VQA bridges a key gap by unifying commonsensse knowledge with LVLM-driven reasoning, eliminating the need for extensive pre-training or complex prompt tuning.
Our framework achieves state-of-the-art performance on benchmark datasets, significantly improving commonsense reasoning in VQA. Our implementation is open-sourced on GitHub\footnote{\url{https://github.com/adlnlp/magic_vqa }} 

\end{abstract}

\section{Introduction}

Visual Question Answering (VQA) \cite{antol2015vqa,goyal2017making, yue2024mmmu} is a complex task requiring models to understand the interaction between visual inputs and textual queries.  In recent years, Large Vision-Language Models (LVLMs) \cite{dai2023instructblipgeneralpurposevisionlanguagemodels, wang2024qwen2, liu2024visual, li2024llava, openai2024gpt4ocard} have made substantial progress in VQA through extensive pre-training on massive image-text datasets and instruction tuning. These models excel at object-level visual recognition and semantic understanding, capturing attributes such as spatial relationships and contextual details.

Nevertheless, LVLMs often face challenges on questions requiring commonsense reasoning, particularly those hinging on implicit contextual cues or everyday world knowledge \cite{zhou2023rome, ye2211improving, li2023vision}\footnote{The sample illustrations can be found in Section \ref{Qualitative_Analysis} and Appendix \ref{additional case studies}.}. To overcome this limitation and improve the performance on , different methods have been explored. For example, Multimodal retrieval-augmented generation leverages dense retrieval to inject external multimodal information into the generation process, thereby enhancing the factual grounding of LVLMs \cite{lin2022retrieval, hu2023reveal}. Multimodal prompt tuning harnesses the model’s innate commonsense knowledge by carefully crafting prompts that combine visual and textual cues from representative samples, guiding LVLMs to leverage their internal reasoning for context-rich answers \cite{wei2022chain, zhang2023multimodal}. However, static prompt design usually lacks the dynamic adaptability required for novel scenarios, resulting in limited generalization to unseen or diverse inputs. Additionally, graph-based approaches utilize Graph Neural Networks (GNNs) to incorporate structured commonsense knowledge \cite{ravi2023vlc,wang2022vqa}, which surpasses the limitations of purely parametric LVLMs, enabling models to capture explicit and implicit knowledge connections via structured graphs. 


However, a key missing component in existing works is the effective integration of commonsense knowledge with LVLMs while addressing their inherent shortcomings. Prior approaches either rely on static retrieval that indiscriminately injects input-unaware noisy knowledge or graph-based augmentation that overlooks the dynamic interplay between external and innate knowledge. Our work seeks to fill this gap by proposing a unified framework that systematically combines dynamic, contextually aligned commonsense integration with structured graph-based reasoning to robustly filter and incorporate relevant commonsense knowledge.

In this paper, we introduce MAGIC-VQA, a novel framework designed to enhance VQA models by effectively integrating commonsense knowledge with LVLMs. MAGIC-VQA is built upon a three-stage process that not only improves reasoning capabilities but also mitigates the complexity associated with large-scale pre-training and inefficient prompt-based approaches.
First, explicit Commonsense Knowledge Integration extracts relevant knowledge triples from external sources, establishing a reliable reasoning foundation. Secondly, by-Type Commonsense Knowledge Post-processing refines these triples based on input-specific needs, ensuring contextual relevance.
Finally, implicit Commonsense Knowledge Augmentation constructs a heterogeneous multimodal graph processed by a GNN to capture intricate relationships, providing structured reasoning beyond what LVLMs alone can infer.
By integrating explicit and implicit commonsense knowledge on top of LVLMs, MAGIC-VQA addresses both the limitations of previous approaches and the missing component in existing works. Our main contributions are as follows:





\begin{enumerate}
    \item We propose MAGIC-VQA, a novel end-to-end framework that systematically integrates both explicit and implicit commonsense knowledge into VQA through, without extensive pre-training or intricate prompt tuning. 
    \item MAGIC-VQA employs a three-stage pipeline—explicit commonsense integration, by-type post-processing, and graph-based implicit augmentation-that dynamically extracts and filters commonsense knowledge in an input-aware manner, and leverages a GNN-based structured reasoning mechanism.
    \item We conduct extensive evaluations across multiple VQA benchmarks, demonstrating robust improvement in commonsense understanding reasoning for VQA, surpassing existing models in both knowledge grounding and inference accuracy.
\end{enumerate}

\section{Related Work}
\subsection{VLPM and LVLMs on VQA}

Vision-Language Pretrained Models (VLPMs) like ViLBERT \cite{su2019vl}, ALBEF \cite{li2021align} and VILT \cite{kim2021viltvisionandlanguagetransformerconvolution} have advanced Visual Question Answering (VQA) by improving the alignment between visual and textual modalities in the last few years\cite{long2022vision}. Recently, Large Vision-Language Models (LVLMs) like InstructBLIP \cite{dai2023instructblipgeneralpurposevisionlanguagemodels}, LLaVA \cite{liu2024visual}, GPT4o \cite{openai2024gpt4ocard} and Gemini1.5 \cite{geminiteam2024gemini15unlockingmultimodal} further push the boundary of VQA with strong in-context learning capability through extensive pre-training and instruction-tuning on large-scale image-text datasets. While these models have demonstrate their effectiveness in domains such as math \cite{jiang2024marvel, wang2025learning}, robotics \cite{10969987}, logical reasoning \cite{bi2025cot},medical \cite{wang2025systematic, wang2025fine} and science \cite{lu2022learn}, these models still face challenges with questions requiring commonsense knowledge that is intuitive and straightforward for humans, such as reasoning based on implicit contextual cues or general world knowledge \cite{ye2023improving, jiang-etal-2023-brainteaser, chen2024commonsense, yang2024multimodal, zhou2025ssfold, bi2025prism, jiang2025hiddendetect}. The resource-intensive nature of these models further makes it infeasible to train a model from scratch for enhanced commonsense understanding \cite{bi2024visual}.

\subsection{Commonsense Knowledge Integration for Visual Question Answering}
Several studies have highlighted the critical role of commonsense knowledge integration in enhancing the performance of VLPMs and LVLMs on VQA tasks \cite{wu2022multi,zhang2022visualcommonsensepretrainedunimodal,  ding2022mukea, wang2024soft,  licore, wang2025vl}.
These methods can be classified into two approaches: explicit commonsense knowledge integration and implicit commonsense knowledge integration.

\textbf{1) Explicit commonsense knowledge integration} directly incorporates external commonsense knowledge into model training through instruction tuning, prompt tuning, or reinforcement learning. For example, VLC-BERT \cite{ravi2023vlc} encodes the contextualized commonsense knowledge of the question phrases as additional textual features and integrates with object visual features to fine-tune the VL-BERT \cite{su2019vl}. MM-CoT \cite{zhang2023multimodal}, T-SciQ \cite{wang2024t} and KAM-CoT \cite{mondal2024kam} fine-tune models on commonsense-augmented Chain-of-Thought (CoT) data to enhance their reasoning processes. Prism \cite{bi2025prism} incorporate the filtered commonsense knowledge into model via dynamic data selection. Vl-Rethinker \cite{wang2025vl} and Pixel Reasoner \cite{su2025pixel} inject commonsense knowledge into vision-language model via reinforcement learning. However, these methods suffer from static commonsense integration without dynamic filtering to adjust to varying input contexts, resulting in potential noise that impedes nuanced reasoning \cite{li2024cogdevelop2k}. 

\textbf{2) Implicit Commonsense Knowledge Integration} focuses on transferring knowledge from a teacher to a student model without directly incorporating external datasets.  
For example, \cite{dai2022enablingmultimodalgenerationclip} distill knowledge from the dual-stream CLIP \cite{radford2021learning} into BART \cite{lewis2019bart}, achieving strong zero-shot performance. \citet{park2024localized} proposed a novel method to distill knowledge from LLMs on specific image regions, then guiding the LLM to infer commonsense knowledge about those areas. However, these methods often overlook the structured interplay among visual, textual, and commonsense cues, limiting their ability to perform nuanced reasoning \cite{wu2025generative}.

\section{MAGIC-VQA}
\label{magic-vqa}

\begin{figure*}[!t]
    \centering
    \includegraphics[width=1\linewidth]{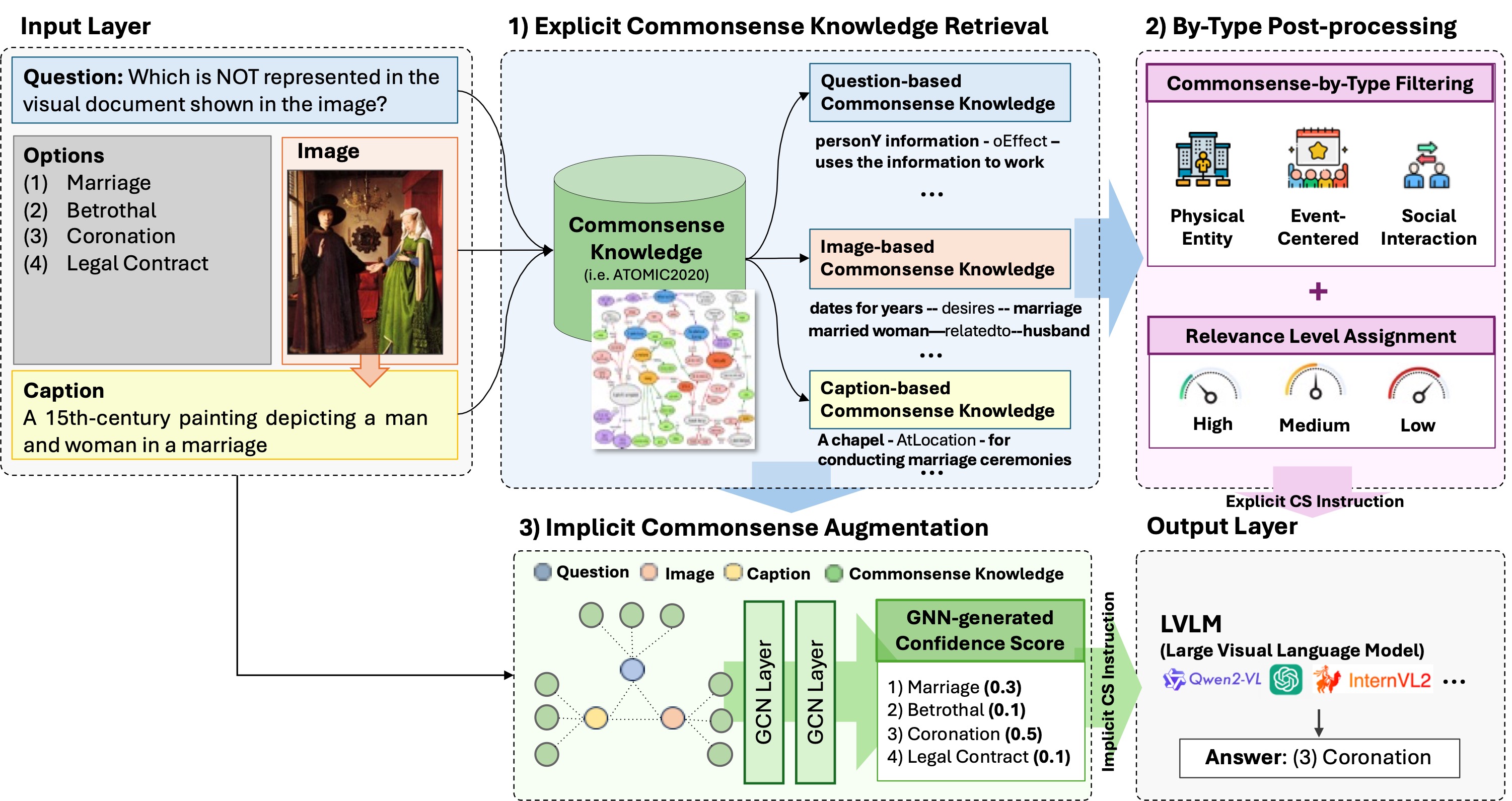}
    \caption{The proposed MAGIC-VQA Framework Architecture, which includes diverse approaches to integrate commonsense knowledge to Visual Question Answering. The detailed description of each step - 1) Explicit Commonsense Knowledge Retrieval, 2) By-Type Post-processing, 3) Implicit Commonsense Augmentation - is aligned with the subsection titles under Section \ref{magic-vqa}.}
    \label{fig:architecture}
\end{figure*}

MAGIC-VQA employs a three-stage process to integrate commonsense knowledge into LVLMs, as in Figure~\ref{fig:architecture}. 
(1) Explicit Commonsense Knowledge Retrieval extracts relevant triples from an external knowledge graph. (2) By-Type Commonsense Knowledge Post-processing refines these triples, aligning them with dataset-specific distributions and assigning relevance levels. (3) Implicit Commonsense Knowledge Augmentation constructs a multimodal graph processed by a GNN to generate confidence scores. These scores, along with the refined triples with relevance level, image, and question, form a comprehensive input to the LVLMs for robust commonsense-grounded inference.

\subsection{Explicit Commonsense Knowledge Integration}
\label{Elementary Commonsense Retrieval}

\begin{table}[!t]
\centering
\renewcommand{\arraystretch}{1.15}
\setlength{\tabcolsep}{7pt}
\begin{adjustbox}{width=\linewidth}
\begin{tabular}{@{}lccc@{}}
\hline
\toprule
\textbf{KG} & \textbf{Size} & \textbf{Main Coverage} & \textbf{Key Relations} \\
\midrule
ConceptNet  & 8M       & PE             & \textit{IsA, UsedFor} \\
ATOMIC      & 877K   & EC, SI                           & \textit{xWant, oEffect}   \\
ATOMIC2020  & 1.33M & PE, EC, SI                       & \textit{23 relation types}        \\
\bottomrule
\hline
\end{tabular}
\end{adjustbox}
\caption{Comparison of three commonsense knowledge graphs. 'PE' refers to physical entity-related commonsense, 'EC' to event-centered related commonsense, and 'SI' to social interaction-related commonsense.}
\vspace{-0.25cm}
\label{tab:kg-comparison}
\end{table}

We begin by integrating explicit commonsense knowledge into LVLM for each input modality.
Given a dataset sample consisting of an image \( I \) and an associated question \( Q \), we first generate an image caption \(C \) using BLIP2 ~\cite{li2023blip} as additional contextual information. Next, we encode the inputs \( \{ I, Q, C \} \) into a shared embedding space, obtaining representations \( f_I \), \( f_Q \), and \( f_C \) using the same model.

We adopt ATOMIC2020~\cite{hwang2021comet} as our external knowledge source because of its broad coverage of physical-entity (PE), event-centered (EC), and social-interaction (SI) relations, as shown in Table~\ref{tab:kg-comparison}. Spanning 1.33 million triplets and 23 relation types, it offers a more balanced scope than either ConceptNet~\cite{speer2017conceptnet} or the earlier ATOMIC~\cite{sap2019atomic}, making it especially relevant for everyday objects, actions, and social contexts encountered in VQA. These 23 relations fall into three groups: (1) \textbf{Physical Entity (PE)}: object properties and functions like “paper is made of cellulose”. (2) \textbf{Event-Centered (EC)}: situational sequences or events, such as “X eats breakfast” typically happening before “X goes to work.” (3) \textbf{Social Interaction (SI)}: human interactions, intentions and emotions, such as “PersonX gives a gift,” leading to “PersonY feels appreciated.” The complete list of relations within each group is covered in Appendix~\ref{commonsense-transformation}.

To retrieve relevant commonsense knowledge,we encode the head and tail entities of all ATOMIC2020 candidates using the same BLIP2 model, then compute cosine similarities between these entity embeddings and input embeddings \( f \in \{ f_I, f_Q, f_C \} \). We select the top \( K \) triplets with the highest cosine similarity scores per input embedding \( f \). This ensures only the contextually pertinent commonsense knowledge is retained, providing a solid foundation for the subsequent refinement and integration stages.

\subsection{By-type Commonsense Knowledge Post-Processing}
\label{Commonsense By-type Post-Processing}

After acquiring an initial pool of commonsense triplets,  
we further refine them through a by-type post-processing stage, ensuring each of them is both tailored to each dataset’s specific needs and contextually aligned. This stage involves two main steps: 
(1) By-type Commonsense Knowledge Filtering, and 
(2) Relevance Level Assignment.

\textbf{By-type Commonsense Knowledge Filtering} customizes the selection of retrieved triplets by matching the desired commonsense type distribution for each dataset. As discovered in Figure~\ref{fig:cs-distribution}, each dataset benefits from a distinct mix of commonsense types. We first discard triplets with similarity scores below a threshold \( \tau \). Let \( T = \{ \text{CS-PE},\ \text{CS-EC},\ \text{CS-SI} \} \) represent the commonsense types, with each type \( t \) allocated a target proportion \( p_t \). We then select \( k_t = \left\lfloor p_t \times k \right\rfloor \) triplets from each type \( t \) with the highest similarity scores, ensuring the final set reflects the dataset’s recommended distribution of commonsense knowledge. Details on these ratios are in Section \ref{sec:Commonsense Type Distribution Analysis}.

\textbf{Relevance Level Assignment} further assign a qualitative relevance level to each filtered triplet based on its cosine similarity score with the input sample, assisting the model in prioritizing most meaningful knowledge during reasoning. For each input source \( f \in \{ f_I, f_Q, f_C \} \), we first aggregate all cosine similarity scores \( S_f = \{ s_j^{(f)} \} \) of the selected triplets. We compute the mean \( \mu_f \) and standard deviation \( \sigma_f \) of these scores for each dataset:
\begin{equation}
\mu_f = \frac{1}{N_f} \sum_{j=1}^{N_f} s_j^{(f)}
\end{equation}
\vspace{-0.2cm}
\begin{equation}
\sigma_f = \sqrt{\frac{1}{N_f} \sum_{j=1}^{N_f} \left( s_j^{(f)} - \mu_f \right)^2}
\end{equation}
where \( N_f \) represents the total number of selected triplets for that input source \( f \). As the scores have a roughly normal distribution, we apply dynamic thresholding that uses mean \( \mu_f \) and standard deviation \( \sigma_f \) to assign each triplet a relevance level:
\begin{equation}
L(s_j^{(f)}) = 
\begin{cases} 
    \textit{High} & \text{if } s_j^{(f)} \geq \mu_f + \frac{\sigma_f}{2} \\[6pt]
    \textit{Medium} & \text{if } \mu_f - \frac{\sigma_f}{2} \leq s_j^{(f)}, \\[4pt]
                    & \text{and } s_j^{(f)} < \mu_f + \frac{\sigma_f}{2} \\[6pt]
    \textit{Low} & \text{if } s_j^{(f)} < \mu_f - \frac{\sigma_f}{2} 
\end{cases}
\end{equation}
Detailed distribution of the similarity score for each dataset is provided in Appendix \ref{fig:Overall caption describing all three distributions}.

\subsection{Implicit Commonsense Knowledge Augmentation}
\label{Commonsense Augmented Confidence}

While explicit retrieval yields relevant commonsense triplets, an implicit augmentation step allows these triplets to be more deeply integrated into the reasoning process. We construct a heterogeneous graph \(G_n = \{V, E\}\) where each input node (image \(I\), question \(Q\), and caption \(C\)) is interconnected and also linked to $k$ additional commonsense nodes. These commonsense nodes are derived by flattening filtered commonsense triplets from Section~\ref{Commonsense By-type Post-Processing}, thereby converting each triplet into a short natural-language sentence for more straightforward integration\footnote{We apply a rule-based triplets flatten mechanism covered in Appendix~\ref{commonsense-transformation}}. Edges between nodes are constructed based on cosine similarity scores between their embeddings, highlighting the semantic relevance between each pair of nodes. The graph is then processed using a two-layer Graph Convolutional Network (GCN) to iteratively update node embeddings: 
\begin{equation}\label{gcn}
     H^{\left ( l+1 \right )} = \rho \left ( \widetilde{A}H^{\left ( l \right )}W_l \right )
\end{equation}
where $\rho$ is a nonlinear activation function and $\widetilde{A}$ is the normalized adjacency matrix. The node embeddings $H^{(2)}$ are pooled to form a unified graph representation for each sample, which is then passed through a Multi-Layer Perceptron (MLP) to produce a confidence score over candidate answers. These confidence scores provide a commonsense-augmented signal to the LVLM, enabling it to prioritize answers grounded in relevant knowledge and improving inference reliability.

\begin{figure}[!t]
    \centering
    \includegraphics[width=1\linewidth]{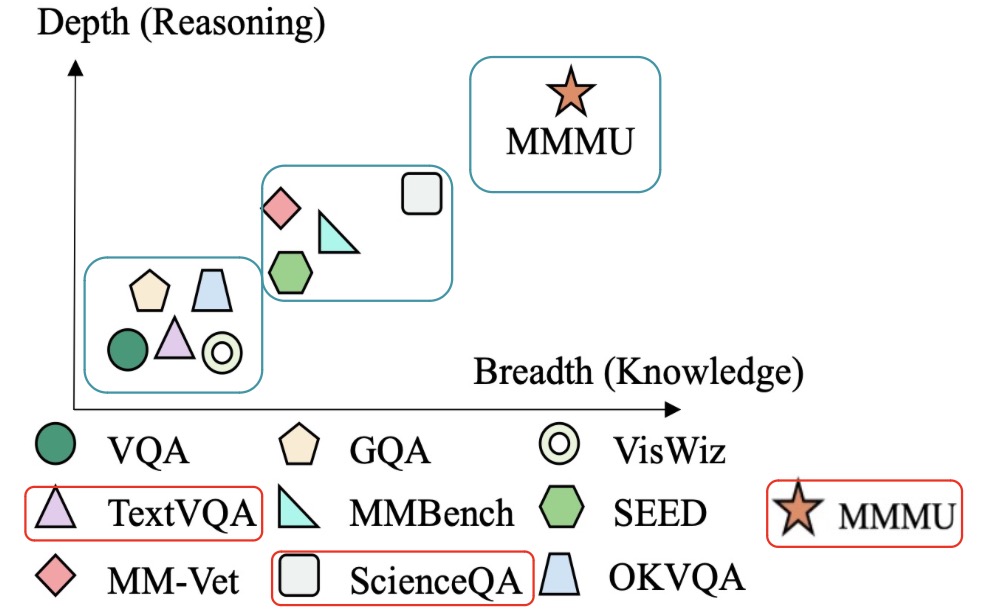}
    \caption{The comparison among VQA datasets. We selected one dataset from each of three groups. We modified the figure from \cite{yue2024mmmu}.}
    \label{fig:Dataset_Depth}
\end{figure}

\subsection{Commonsense Grounded Inference}
\label{Commonsense Grounded Inference}

In the final inference stage, we combine all processed elements—original inputs (\(I, Q, C\)), refined commonsense triplets (with assigned relevance levels) and GNN-generated confidence scores—into a unified input structure for inference with LVLMs. A complete input example is provided in Appendix~\ref{Concrete Example of Input Prompt}. By fusing explicit and implicit commonsense knowledge with visual and textual signals, the LVLMs can reason effectively about nuanced relationships, delivering answers better aligned with real-world understanding.

\section{Experiment}

\subsection{Dataset}

We evaluated MAGIC-VQA on three representative VQA benchmarks of diverse complexity and depth as highlighted in Figure~\ref{fig:Dataset_Depth}. 

\textbf{ScienceQA} \cite{lu2022learn} comprises over 21,000 multiple-choice questions from elementary and middle school curricula in natural, social, and language science. It tests factual and procedural understanding, requiring integration of commonsense about the physical world and scientific phenomena. We select only samples with image contexts.

\textbf{TextVQA} \cite{singh2019towards} contains over 45,000 questions grounded in 28,000 real-world images with embedded text like signs and labels. It demands OCR to extract textual elements and integrate them with everyday commonsense provided in the context to interpret them within the visual scene. We use its validation set for our evaluation.

\textbf{MMMU} \cite{yue2024mmmu} consists of 11,550 college-level questions spanning diverse disciplines. It features challenging  image types such as medical diagnosis, music sheets and so on, which goes beyond the everyday commonsense understanding emphasized in ScienceQA and TextVQA.  We choose its validation set to evaluate our model.

\begin{figure}[!t]
    \centering
    \includegraphics[width=0.85\linewidth]{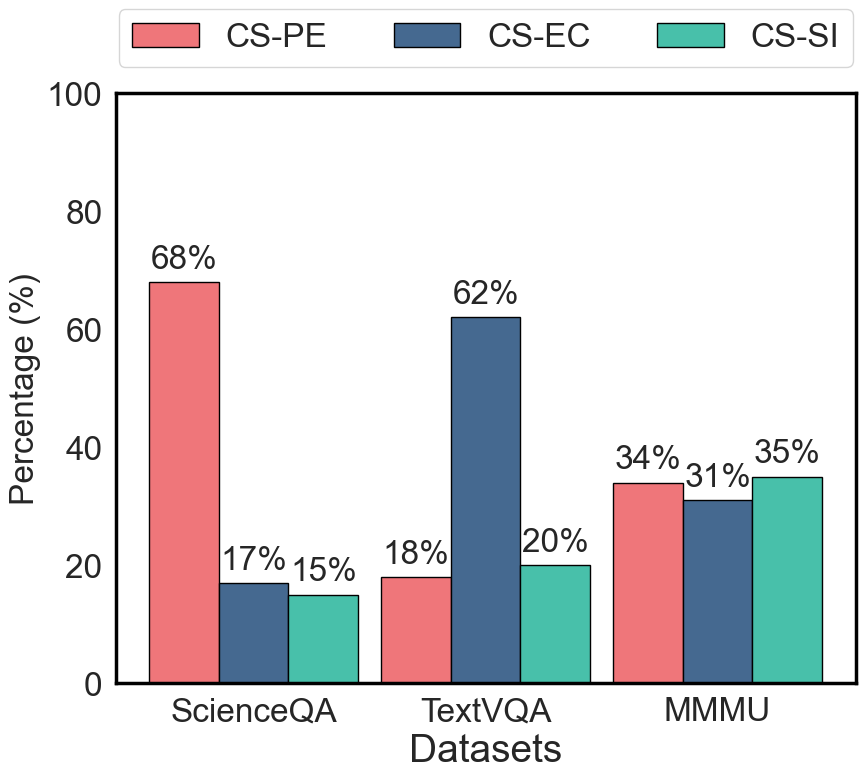}
    \caption{The distribution of categories of commonsense. CS-PE refers to physical entity-related commonsense, CS-EC to event-centered related commonsense, and CS-SI to social interaction-related commonsense.}
    \label{fig:cs-distribution}
\end{figure}

\subsection{Commonsense Knowledge Distribution}
\label{sec:Commonsense Type Distribution Analysis}

\begin{table*}[htbp]
    \centering
    \renewcommand{\arraystretch}{1.1} 
    \caption{Performance comparison under four configurations:
(1) \textit{None}: Inputs with no additional commonsense knowledge.
(2) \textit{CS Sources}: Inputs enriched with commonsense knowledge from different sources. including question (CS-Q), image (CS-I), and image caption (CS-C).
(3) \textit{CS Categories}: Inputs enriched with commonsense knowledge from different categories, including Physical Entities (CS-PE), Event-Centered (CS-EC), and Social Interaction (CS-SI).
(4) \textit{All CS}: Inputs enriched with all source-based and category-based commonsense.}
    \setlength{\tabcolsep}{10pt} 
    \small 
    \definecolor{verylightgray}{gray}{0.9} 
    \begin{adjustbox}{width=\linewidth} 
    \begin{tabular}{c|c|ccc|ccc|c}
        \hline
        \specialrule{1.5pt}{0pt}{0pt}
        \multirow{2}{*}{\textbf{Models}} & \multirow{2}{*}{\textbf{None}} & \multicolumn{3}{c|}{\textbf{CS Sources}} & \multicolumn{3}{c|}{\textbf{CS Categories}} & \multirow{2}{*}{\textbf{All CS}} \\
        & & \textbf{CS-Q} & \textbf{CS-I} & \textbf{CS-C} & \textbf{CS-PE} & \textbf{CS-EC} & \textbf{CS-SI} & \\
        \hline
        \rowcolor{verylightgray} \multicolumn{9}{c}{\cellcolor{verylightgray} \textbf{ScienceQA\(_{IMG}\)}} \\ \hline
        \textbf{LLaVA1.6} & 67.50 & 68.83 & \underline{71.56} & 70.35 & \underline{71.12} & 69.01 & 70.83 & \textbf{72.30} \\
        \textbf{BLIP3} & 70.00 & 71.56 & \underline{73.88} & 72.97 & \underline{74.03} & 71.57 & 71.05 & \textbf{74.30} \\
        \textbf{InternVL2} & 71.99 & 72.58 & \underline{74.37} & 73.91 & \underline{74.56} & 73.09 & 73.21 & \textbf{74.62} \\
        \textbf{Qwen2VL} & 71.39 & 72.21 & \underline{74.83} & 71.86 & \underline{74.22} & 72.03 & 72.57 & \textbf{75.95} \\
        \textbf{GPT4o-mini} & 76.45 & 77.34 & \underline{79.83} & 77.17 & \underline{79.63} & 77.52 & 78.87 & \textbf{81.22} \\
        \hline
        \rowcolor{verylightgray} \multicolumn{9}{c}{\cellcolor{verylightgray} \textbf{TextVQA\(_{val}\)}} \\ \hline
        \textbf{LLaVA1.6} & 62.30 & 63.55 & \underline{64.82} & 64.23 & 64.77 & \underline{65.05} & 64.89 & \textbf{65.20} \\
        \textbf{BLIP3} & 67.80 & 68.49 & \underline{69.64} & 68.29 & 69.12 & \underline{69.64} & 69.24 & \textbf{69.80} \\
        \textbf{InternVL2} & 73.21 & 74.06 & \underline{75.19} & 74.81 & 74.60 & \underline{75.01} & 74.82 & \textbf{75.30} \\
        \textbf{Qwen2VL} & 75.30 & 76.07 & \underline{77.63} & 77.05 & 76.57 & \underline{78.02} & 76.85 & \textbf{78.90} \\
        \textbf{GPT4o-mini} & 78.98 & 79.34 & \underline{81.25} & 80.63 & 80.93 & \underline{81.51} & 81.22 & \textbf{82.13} \\
        \hline
        \rowcolor{verylightgray} \multicolumn{9}{c}{\cellcolor{verylightgray} \textbf{MMMU\(_{val}\)}} \\ \hline
        \textbf{LLaVA1.6} & 48.38 & 49.27 & \underline{53.52} & 49.85 & 52.03 & 52.57 & \underline{53.10} & \textbf{54.30} \\
        \textbf{BLIP3} & 41.31 & 42.54 & \underline{45.89} & 42.19 & 44.12 & \underline{46.03} & 45.89 & \textbf{47.60} \\
        \textbf{InternVL2} & 51.00 & 52.17 & \underline{55.48} & 54.21 & \underline{54.23} & 52.67 & 53.50 & \textbf{55.80} \\
        \textbf{Qwen2VL} & 51.10 & 52.69 & \underline{55.89} & 54.83 & 53.60 & \underline{54.57} & 54.10 & \textbf{57.42} \\
        \textbf{GPT4o-mini} & 55.89 & 56.53 &         
        \underline{58.79} & 56.21 & \underline{58.12} & 57.57 & 57.89 & \textbf{60.87} \\
        \specialrule{1.5pt}{0pt}{0pt}
    \end{tabular}
    \end{adjustbox}
    \label{tab:Unified-results}
\end{table*}

To tailor the commonsense knowledge to each dataset’s specific reasoning requirements, we analyze the distribution of commonsense types across each dataset using GPT4 \cite{openai2024gpt4technicalreport}\footnote{The prompt template is in Appendix \ref{Commonsense Category Analysis Prompt Format}}. As Figure~\ref{fig:cs-distribution} suggests,  ScienceQA requires more Physical Entity (CS-PE) knowledge, possibly due to its focus on factual scientific concepts. Meanwhile, TextVQA, which often involves contextual understanding in images, benefits more from Event-Centered (CS-EC) knowledge. MMMU, however, requires a balanced mix of Physical Entity, Event-Centered, and Social Interaction (CS-SI) commonsense due to its multi-disciplinary nature.  As a result, we set the by-type filtering ratio of \{CS-PE:CS-EC:CS-SI\} mentioned in Section~\ref{Commonsense By-type Post-Processing} as \{0.7:0.15:0.15\} in ScienceQA, \{0.2:0.6:0.2\} in TextVQA,\{0.33:0.33:0.33\} in MMMU dataset.

\subsection{Baselines, Metric, and Implementations}
The selected baselines are four open source state-of-the-art LVLMs: LLaVA-1.6 \cite{liu2024llava}, XGen-MM (BLIP-3) \cite{xue2024xgen}, InternVL2 \cite{chen2024far}, Qwen2VL \cite{wang2024qwen2}, and one proprietary model, GPT4o-mini \cite{openai2024gpt4ocard}. These LVLMs are selected for their outstanding zero-shot performance in VQA tasks. Details of each baseline model are in Appendix \ref{Baseline details}. We adopt accuracy as the evaluation metric following prior works \cite{singh2019towards, lu2022learn, yue2024mmmu}. All experiments are conducted with and without the proposed MAGIC-VQA under a zero-shot setup. More implementation details are in Appendix \ref{implementation details}.

\section{Results}
\subsection{Explicit Commonsense Knowledge}
\label{ex-cs}

We evaluated the explicit integration of commonsense knowledge triplets by systematically testing four configurations of: (1) \textit{None}: Inputs with no additional commonsense; (2) \textit{CS Sources}: Inputs augmented with commonsense from questions, images, or captions; (3) \textit{CS Categories}: Inputs augmented with commonsense grouped by category (Physical Entities, Event-Centered, Social Interaction); and (4) \textit{All CS}: Inputs augmented with all retrieved commonsense\footnote{Each experiment is tested with a fixed number of $k = 6$ to maintain a fair comparison. We further investigated the impact of incorporating different numbers of commonsense knowledge triplets in Appendix~\ref{num_of_triplet}.}. As in Table~\ref{tab:Unified-results}, integrating explicit commonsense consistently improves performance across all baselines and three datasets. For instance, on ScienceQA, GPT-4O’s accuracy rises from 76.45\% (\textit{None}) to 81.22\% (\textit{All CS}), and Qwen2VL improves from 51.10\% to 57.42\% on MMMU under the same setup.
Examining the effect of source-based commonsense reveals that image-driven knowledge (CS-I) typically provides the largest gains. For example, LLaVA1.6 on MMMU jumps from 48.38\% to 53.52\% with CS-I, surpassing the minor improvements from CS-Q or CS-C. This suggests that leveraging image-aligned commonsense offers more grounded cues for inference. However, category-based commonsense (CS-PE, CS-EC, and CS-SI) exhibits dataset-dependent effectiveness. On ScienceQA, CS-PE is most beneficial, while on TextVQA, CS-EC dominates, and MMMU shows a more balanced pattern. These results align with our earlier commonsense distribution analysis in Section~\ref{sec:Commonsense Type Distribution Analysis}, highlighting the importance of tailoring knowledge retrieval to the dataset’s unique characteristics.

\subsection{Implicit Commonsense Knowledge}

\begin{table}[!t]
\centering
\renewcommand{\arraystretch}{1.3}
\setlength{\tabcolsep}{1.5pt}
\begin{adjustbox}{width=\linewidth}
\begin{tabular}{l|cc|c|ccc} 
\hline
\specialrule{2pt}{0pt}{0pt}
\Large
\multirow{2}{*}{\textbf{Model}} & \multicolumn{2}{c|}{\textbf{Ex-CS}} & \textbf{Im-CS} & \multicolumn{3}{c}{\textbf{Performance}} \\ 
\cmidrule(lr){2-3} \cmidrule(lr){4-4} \cmidrule(lr){5-7}
 & \textbf{CS} & \textbf{Rel} & \textbf{Conf} & \textbf{SQA} & \textbf{MMMU} & \textbf{TVQA} \\ 
\midrule
Qwen2VL      & \xmark & \xmark & \xmark  & 71.39 & 51.10 & 75.30 \\
Qwen2VL      & \cmark & \xmark & \xmark  & 75.11 & 56.00 & 78.50 \\ 
Qwen2VL      & \xmark & \xmark & \cmark  & 72.88 & 53.41 & 76.42 \\
Qwen2VL      & \cmark & \cmark & \xmark  & 75.95 & 57.42 & 78.90 \\
Qwen2VL      & \cmark & \xmark & \cmark  & 76.42 & 57.21 & 79.10 \\ 
\textbf{Qwen2VL} & \textbf{\cmark} & \textbf{\cmark} & \textbf{\cmark} & \textbf{77.12} & \textbf{58.72} & \textbf{79.80} \\ 
\midrule
GPT4o-mini        & \xmark & \xmark & \xmark & 76.45 & 55.89 & 78.98 \\ 
GPT4o-mini        & \cmark & \xmark & \xmark & 80.07 & 59.30 & 81.73 \\
GPT4o-mini        & \xmark & \xmark & \cmark & 77.02 & 57.64 & 79.55 \\ 
GPT4o-mini        & \cmark & \cmark & \xmark & 81.22 & 60.87 & 82.13 \\
GPT4o-mini        & \cmark & \xmark & \cmark & 80.94 & 60.25 & 82.50 \\ 
\textbf{GPT4o-mini} & \textbf{\cmark} & \textbf{\cmark} & \textbf{\cmark} & \textbf{82.50} & \textbf{61.03} & \textbf{83.37} \\ 
\hline
\specialrule{1.5pt}{0pt}{0pt}
\end{tabular}
\end{adjustbox}
\caption{Quantitative analysis on the effect of each component of MAGIC-VQA. The "Ex-CS" ($CS$ and $Rel$) denotes explicit commonsense knowledge inclusion (All-CS in Section \ref{ex-cs}), while "Im-CS" ( $Conf$) denote implicit commonsense inclusion. (\cmark) and (\xmark)  denotes the inclusion and exclusion of a component.}
\label{tab:LVLM-component_analysis}
\end{table}

We next examined the effect of implicit commonsense knowledge augmentation as outlined in Section~\ref{Commonsense Augmented Confidence} using two representative LVLMs, Qwen2VL and GPT4o-mini. We also include results from explicit commonsense knowledge integration (All-CS in Section \ref{ex-cs}) as a reference.  As Table~\ref{tab:LVLM-component_analysis} suggests, while the implicit commonsense knowledge $Conf$ does not contribute as significantly as explicit commonsense knowledge (\textit{Ex-CS}), it nonetheless provides complementary information that enhances overall performance. Incorporating only $Conf$ with Qwen2VL improves the MMMU accuracy from 51.10\% to 53.41\%. We also observe that adding $Rel$ notably improves results across all three datasets. Furthermore, combining implicit and explicit commonsense yields the highest overall performance, indicating that implicit augmentation complements explicit knowledge by capturing additional context that explicit methods alone may miss. Further analysis is provided in Section \ref{Qualitative_Analysis} and Appendix \ref{additional case studies}. 

\begin{table}[!t]
\centering
{%
  \huge                          
  \renewcommand{\arraystretch}{1.2}
  \begin{adjustbox}{width=\linewidth}
    \begin{tabular}{@{} l l S[table-format=2.2, detect-weight] S[table-format=2.2, detect-weight] @{}}
    \specialrule{2pt}{0pt}{0pt}
    \textbf{Model} & \textbf{Knowledge} & {\textbf{A-OKVQA}} & {\textbf{VCR}} \\
    \midrule
    VLCBERT & COMET & 38.05 & 79.24 \\
    KAT & Wikidata + GPT3 & 49.74 & 83.18 \\
    KRISP & Wikidata + CNet & 27.10 & 65.12 \\
    \hdashline
    Ours (LLaVA1.6) & ATOMIC2020 & 73.33 & 89.91 \\
    Ours (Qwen2VL) & ATOMIC2020 & 75.95 & 91.24 \\
    Ours (GPT4o-mini) & ATOMIC2020 & \textbf{76.55} & \textbf{93.42} \\
    \specialrule{2pt}{0pt}{0pt}
    \end{tabular}
  \end{adjustbox}%
}%
\caption{Comparison of knowledge-augmented VQA models on two benchmarks. CNet stands for the Conceptnet dataset}
\label{tab:knowledge_vqa}
\end{table}

\subsection{Comparison With Other Baselines}

To further assess the effectiveness of MAGIC-VQA’s knowledge injection, we compare it with three representative knowledge-augmented baselines including KAT \cite{gui2021kat}, VLC-BERT \cite{ravi2023vlc} and KRISP \cite{marino2021krisp} on two knowledge-intensive VQA benchmarks: VCR \cite{zellers2019recognitioncognitionvisualcommonsense} and the multiple-choice questions subset of A-OKVQA \cite{schwenk2022okvqa}.
As shown in Table \ref{tab:knowledge_vqa}, MAGIC-VQA consistently outperforms all competitors on both datasets. In particular, the GPT-4o-mini variant achieves the highest accuracy of 76.55\% on A-OKVQA and 93.42\% on VCR, exceeding KAT by more than 26\% and 10\%, respectively.
This improvement highlight the strength of our commonsense-infused visual reasoning. Moreover, unlike prior knowledge-augmented models that require specialized pre-training, MAGIC-VQA is a plug-and-play framework that can be seamlessly integrated with different backbones, making it easily extensible to future models and knowledge bases.

\subsection{Ratio of Commonsense Knowledge Type}
\label{ratio_ckt}
\begin{figure}[!t]
    \centering
    \includegraphics[width=1\linewidth]{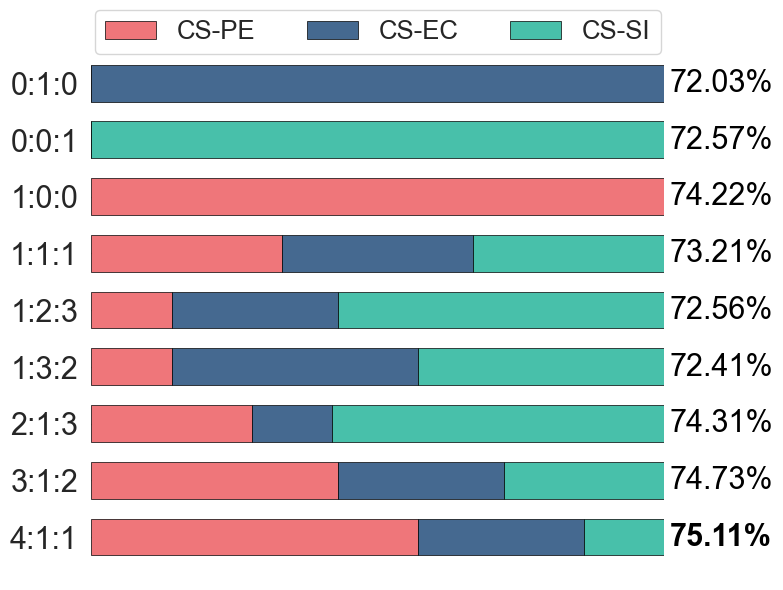}
    \caption{Effect of the ratio of commonsense knowledge categories on Qwen2VL. The three categories of explicit commonsense knowledge include social interactions (CS-SI), physical entities (CS-PE), and event-based relations (CS-EC).}
    \label{fig:result_ratio_cs}
\end{figure}

To determine the optimal ratio of commonsense knowledge types, we conduct an experiment to analyze how varying distributions of CS-PE, CS-EC, and CS-SI knowledge triplets impact the performance of Qwen2-VL on the ScienceQA dataset. Figure \ref{fig:result_ratio_cs} indicates that the model achieves its highest accuracy of 75.11\% with a 4:1:1 ratio, which places greater emphasis on CS-PE knowledge. Additionally, distributions favoring CS-PE triplets consistently result in improved performance. For example, a 3:1:2 ratio achieves an accuracy of 74.73\%. In contrast, ratios prioritizing CS-EC or CS-SI, such as 1:2:3 or 1:3:2, yield lower accuracies of 72.56\% and 72.41\%, respectively. These results suggest that CS-PE is the most essential commonsense knowledge type for the ScienceQA, aligning with its focus on physical concepts and entities as discussed in Section \ref{ex-cs}. Notably, the 4:1:1 ratio closely mirrors the inherent distribution of commonsense knowledge in ScienceQA in Figure \ref{fig:cs-distribution}. This alignment suggests tailoring the balance of commonsense knowledge to match the dataset’s inherent characteristics, as demonstrated by our approach, leads to the most significant performance improvements.  We further attach the effect of diverse similarity metrics by comparing Manhattan, Cosine, and Euclidean Distance when applied to the explicit knowledge retrieval in Appendix~\ref{effect_retrival}.

\subsection{Break Down Results}

\begin{figure}[!t]
    \centering
    \begin{subfigure}[htbp]{\linewidth}
        \centering
        \includegraphics[width=\linewidth]{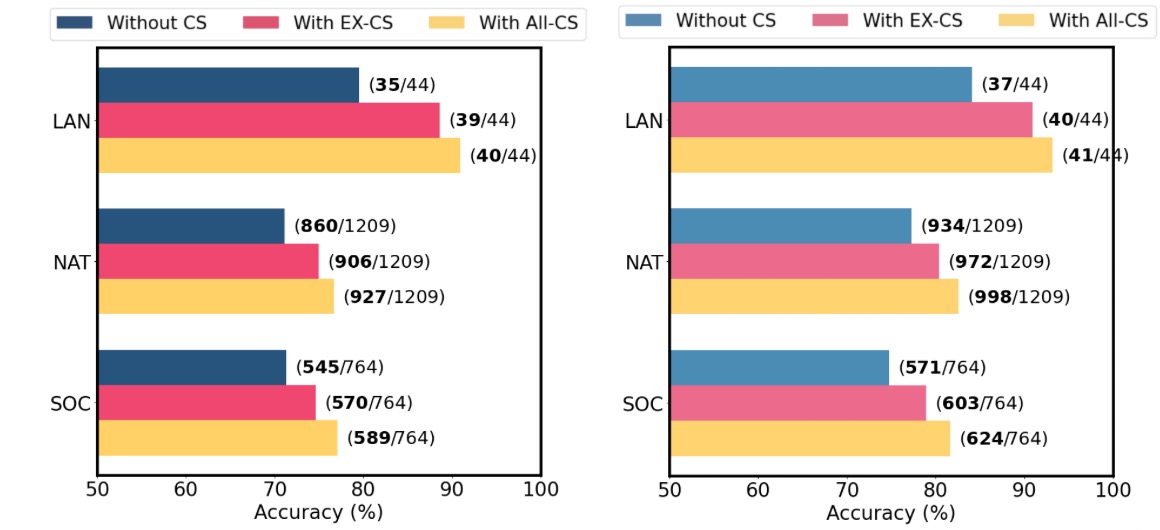}
        \vspace{-0.5cm}
        \caption{ScienceQA}
        \label{fig:scienceqa-break-down}
    \end{subfigure}

    \begin{subfigure}[htbp]{\linewidth}
        \centering
        \includegraphics[width=\linewidth]{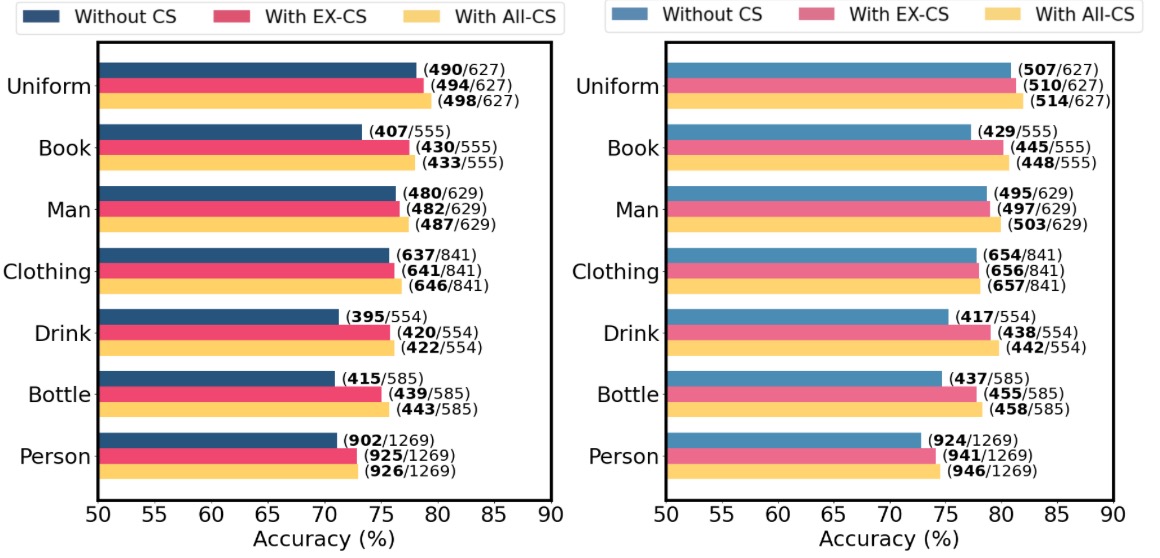}
        \vspace{-0.5cm}
        \caption{TextVQA}
        \label{fig:textvqa-break-down}
    \end{subfigure}

    \begin{subfigure}[htbp]{\linewidth}
        \centering
        \includegraphics[width=\linewidth]{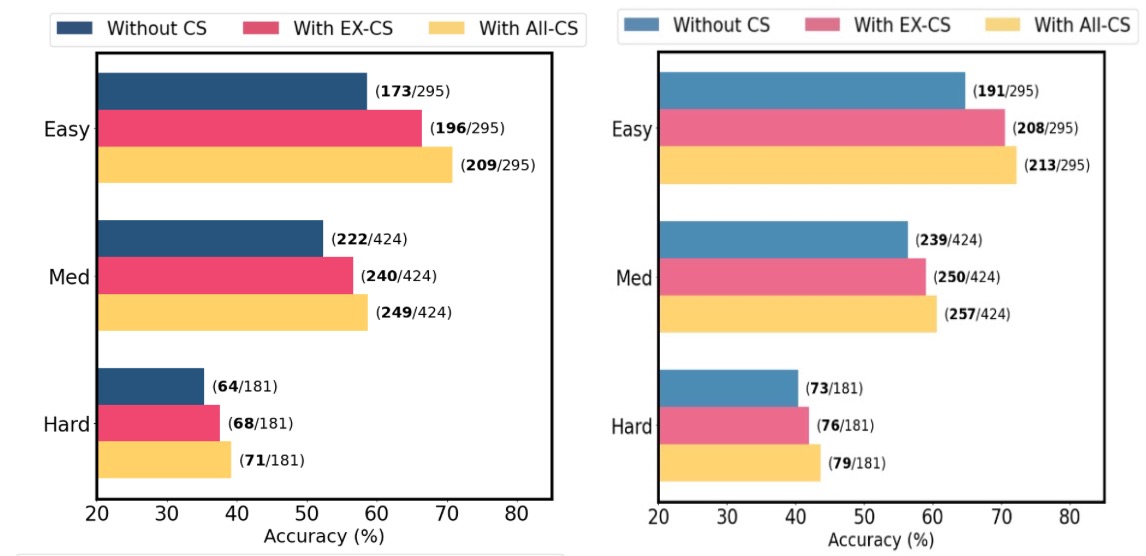}
                \vspace{-0.5cm}
        \caption{MMMU}
        \label{fig:mmmu-break-down}
    \end{subfigure}

    \caption{Subcategory-level accuracy on (a) ScienceQA, (b) TextVQA, and (c) MMMU for Qwen2VL (left) and GPT4o-mini (right) under three conditions: Without CS, With EX-CS and With All-CS.}
    \label{fig:merged-vertical-figure}
\end{figure}

We provide break down result on subcategories within each selected dataset using Qwen2VL and GPT4o-mini as illustrated in Figure~\ref{fig:merged-vertical-figure}. Across all datasets and subcategories, incorporating commonsense significantly improves the accuracy. Each dataset features distinct subcategories that would benefit from varying aspects of commonsense reasoning. First, as illustration of ScienceQA in Figure~\ref{fig:scienceqa-break-down}, commonsense augmentation yields notable improvements in language-related subcategories, reflecting the value of context-sensitive reasoning. Next shown in Figure~\ref{fig:textvqa-break-down} for TextVQA, categories involving concrete objects such as \textit{`uniform'} and \textit{`books'}, benefit more from commonsense augmentation compared to abstract categories like \textit{`persons'}, indicating concrete objects allow for more precise retrieval of relevant knowledge. MMMU in Figure~\ref{fig:mmmu-break-down}, commonsense augmentation benefits \textit{easy}-level questions closely tied to everyday knowledge, while struggles with \textit{hard}-levels that demand complex reasoning beyond commonsense.

\subsection{Computation Resource Comparison}

We provide a computational resources analysis to demonstrate MAGIC-VQA’s exceptional efficiency and scalability. As highlighted in Table \ref{tab:model-comparison-efficiency}, our entire framework spans only 0.33 M trainable parameters, vastly leaner than conventional vision–language transformers, knowledge‐augmented transformers, multimodal chain-of-thought models, and today’s large VLMs.
This dramatic reduction stems from MAGIC-VQA’s plug-and-play architecture: explicit knowledge retrieval and rule-based filtering are parameter-free, while the implicit commonsense module trains only a lightweight GNN. Unlike previous work that must pre-train or fine-tune hefty backbones to absorb external knowledge, our design cleanly decouples knowledge acquisition from model capacity, enabling rapid adaptation to new vision–language encoders with negligible computational overhead.

\begin{table}[!t]
  \centering
  \scriptsize
  \begin{adjustbox}{width=\linewidth}
    \begin{tabular}{lll}
      \toprule
      \textbf{Category / Model} & \textbf{Year} & \textbf{Size} \\
      \midrule
      \rowcolor{gray!20}\textbf{Vision–Language Transformers} & & \\
      \hspace{2em}VL-BERT \tcite{su2019vl}               & 2019 & 111M \\
      \hspace{2em}ViLT \tcite{kim2021viltvisionandlanguagetransformerconvolution} & 2021 & 112M \\
      \hspace{2em}UnifiedQA \tcite{khashabi2020unifiedqa} & 2020 & 223M \\
      \rowcolor{gray!20}\textbf{Knowledge Augmented Transformers} & & \\
      \hspace{2em}KRISP \tcite{marino2021krisp}          & 2020 & 116M \\
      \hspace{2em}KAT \tcite{gui2021kat}                 & 2021 & 175B \\
      \hspace{2em}VICBERT \tcite{ravi2023vlc}            & 2022 & 118M \\
      \rowcolor{gray!20}\textbf{Multimodal CoT Models} & & \\
      \hspace{2em}MM-COT \tcite{zhang2023multimodal}     & 2023 & 223M \\
      \hspace{2em}KAM-COT \tcite{mondal2024kam}          & 2024 & 250M \\
      \rowcolor{gray!20}\textbf{Large Vision-Language Models} & & \\
      \hspace{2em}LLaVA \tcite{liu2023visual}            & 2023 & 7B    \\
      \hspace{2em}GPT4 \tcite{openai2024gpt4technicalreport} & 2024 & >175B \\
      \hspace{2em}Qwen2-VL \tcite{wang2024qwen2}         & 2024 & 7B    \\
      \rowcolor{gray!20}\textbf{Ours} & & \\
      \hspace{2em}\textbf{MAGIC-VQA} & 2024 & \textbf{0.33M} \\
      \bottomrule
    \end{tabular}
  \end{adjustbox}
 \caption{Comparison with notable previous methods from different categories with respect to parameter size.}
 \label{tab:model-comparison-efficiency}
\end{table}

\section{Qualitative Analysis}
\label{Qualitative_Analysis}

\begin{figure}[t!]
    \centering
    \begin{subfigure}{0.51\linewidth}
        \centering
        \includegraphics[width=\textwidth]{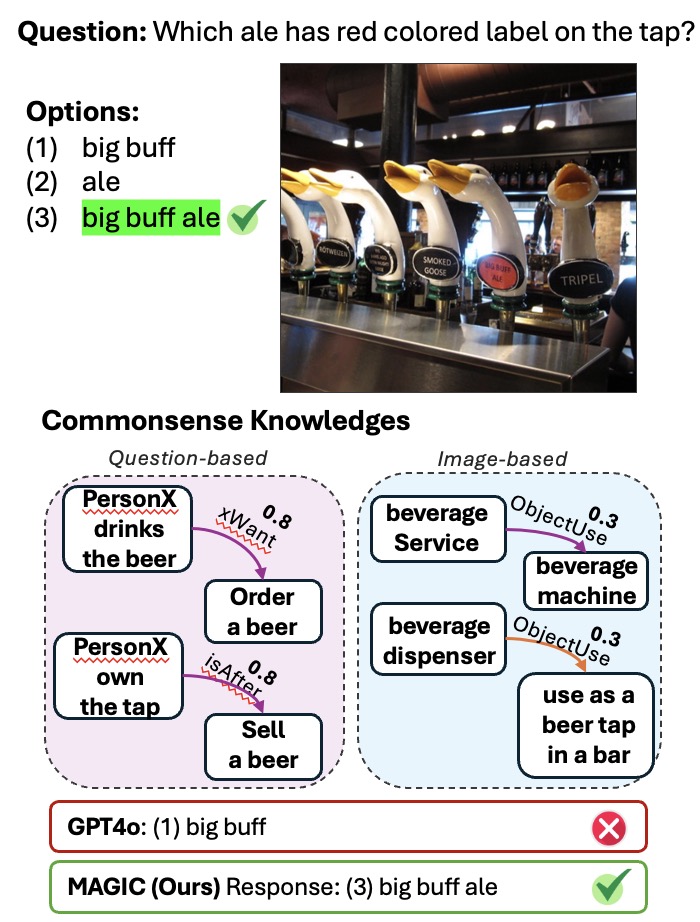}
        \caption{TextVQA example}
        \label{fig:case-study-1}
    \end{subfigure}
    \hfill
    \begin{subfigure}{0.47\linewidth}
        \centering
        \includegraphics[width=\textwidth]{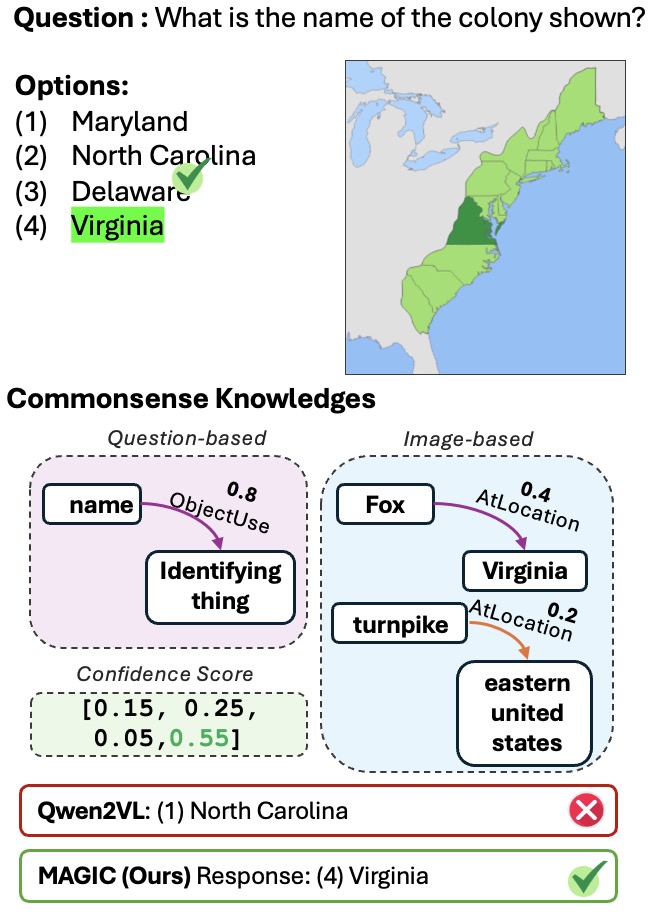}
        \caption{ScienceQA example}
        \label{fig:case-study-2}
    \end{subfigure}

    \caption{Comparison of results of commonsense knowledge-injected MAGIC-VQA (ours) and original GPT4o-mini and Qwen2VL across different datasets, including TextVQA and ScienceQA. Each example highlights the question-based and image-based explicit commonsense knowledge. Example in ScienceQA is also injected with implicit commonsense knowledge.}
    \label{fig:Case_Study}
\end{figure}

Figure \ref{fig:Case_Study} compares MAGIC-VQA with GPT4o and Qwen2VL, demonstrating how our framework effectively integrates both explicit and implicit commonsense knowledge for enhanced visual question answering. As illustrated in Figure \ref{fig:case-study-1}, while GPT4o struggles to deduce the complete answer (big buff ale) to input query, 
MAGIC-VQA successfully incorporate contextual knowledge, such as “Person X owns the tap sells beer” and “beverage dispenser used as a beer tap in a bar,” linking beer consumption, tap functionality, and beverage machines with the input question to arrive at the correct answer. In Figure \ref{fig:case-study-2}, while Qwen2VL incorrectly identifies the colony as North Carolina, our MAGIC-VQA addresses this limitation by integrating explicit image-based commonsense knowledge about Virginia’s location, historical turnpikes, and wildlife, correctly concluding that the answer is Virginia.  Confidence scores derived from implicit commonsense knowledge further reinforce the evidence for the final accurate prediction. Additional case studies demonstrating the characteristics of each dataset and the benefit from each commonsense knowledge injection component are shown in Appendix \ref{additional case studies}.

\section{Conclusion}
This paper introduced MAGIC-VQA, a novel framework integrating commonsense knowledge into VQA to address the limitations of existing LVLMs. MAGIC-VQA's three-stage process—knowledge retrieval, commonsense post-processing, and GNN-based augmentation—enables nuanced reasoning without extensive pre-training or complex prompt tuning. Evaluations on ScienceQA, TextVQA, and MMMU demonstrate significant improvements in tasks requiring advanced reasoning. This framework establishes a robust approach for bridging raw visual inputs with high-level reasoning, offering scalable enhancements for VQA. We hope this work inspires further research into structured commonsense reasoning for complex multimodal challenges.

\section*{Limitation}

While the MAGIC-VQA framework demonstrates significant improvement, it currently relies on external knowledge graphs, such as ATOMIC2020 and predefined commonsense categories, which may limit its adaptability to diverse and unforeseen domains. Additionally, real-world VQA scenarios often involve noisy or ambiguous inputs that may not always align with the structured assumption of the commonsense knowledge graph.
To address these limitations, we plan to extend our approach by developing and incorporating a more diverse and extensive range of multimodal commonsense knowledge sources. Expanding the scope of knowledge representation will enhance multimodal understanding and learning ability and help us handle more multimodal reasoning tasks.

\section*{Acknowledgment}
This study was supported by funding from the Google Award for Inclusion Research Program (G222897).

\bibliography{custom}
\clearpage

\appendix

\section{Additional Experiment Results}
\label{additional experiment Results}


\subsection{Number of Knowledge Triplets}
\label{num_of_triplet}

\begin{figure}[!htbp]
    \centering
    \begin{subfigure}[b]{0.49\linewidth}
        \centering
        \includegraphics[width=\linewidth, height = \linewidth]{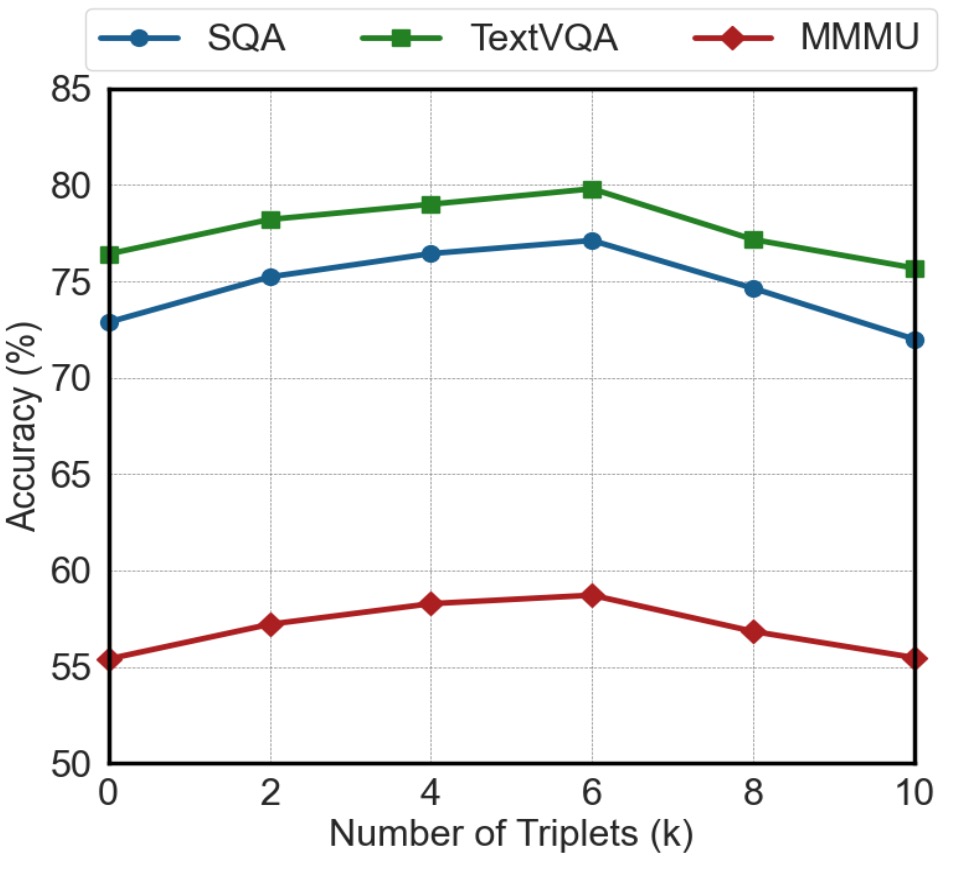}
        \caption{Qwen2VL}
        \label{fig:effect-triplet-number-qwen}
    \end{subfigure}
    \begin{subfigure}[b]{0.49\linewidth}
        \centering
        \includegraphics[width=\linewidth, height = 1.02\linewidth]{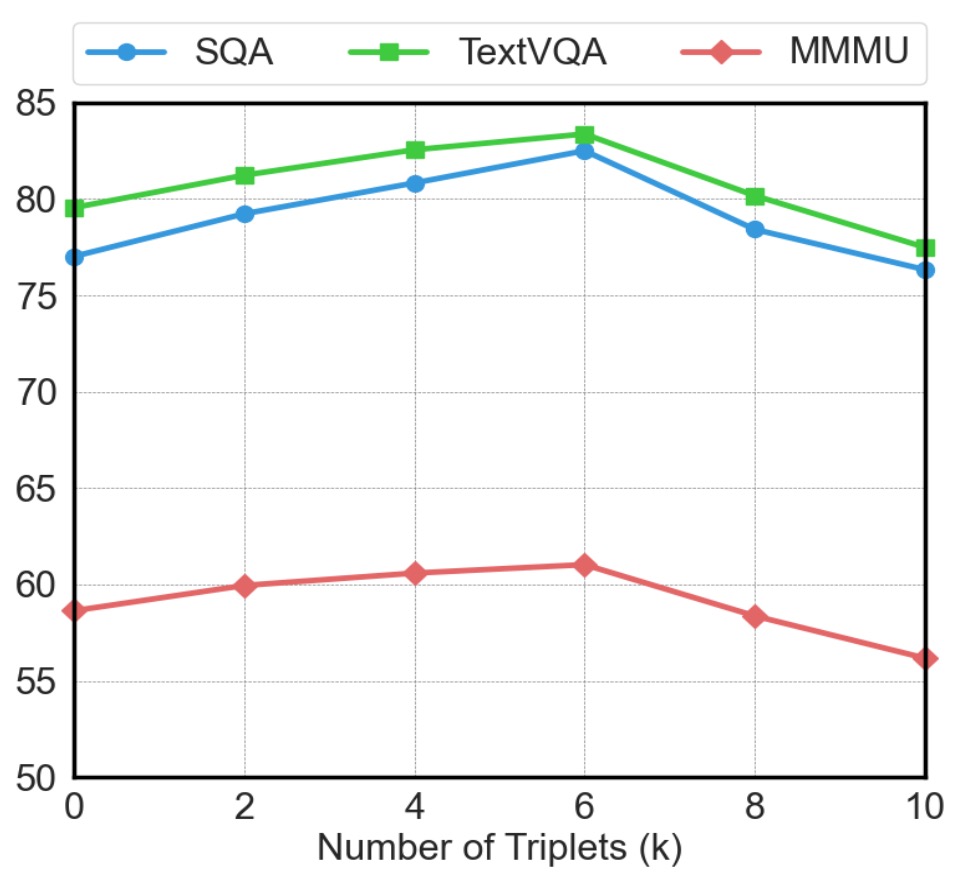}
        \caption{GPT4o-mini}
        \label{fig:effect-triplet-number-gpt}
    \end{subfigure}
    \caption{Effect of number of knowledge triplets $k$: Comparison between Qwen2VL andGPT4o-mini.}
    \label{fig:combined-effect-triplet-number}
\end{figure}

Inspired by \cite{wu2023image}, to optimize the integration of explicit commonsense knowledge into our model while minimizing the risk of introducing excessive noise, we investigate how varying the number of retrieved commonsense triplets \(k\) impacts performance. We vary \(k\) from 0 to 10, incrementally increasing the number of triplets provided to Qwen2VL andGPT4o-mini and measuring the corresponding accuracy.
In Figure \ref{fig:combined-effect-triplet-number}, the models achieve optimal performance when \(k = 6\) triplets are incorporated. Utilizing fewer than \(k = 6\) appears insufficient to provide the contextual information for efficient reasoning, while exceeding \(k = 6\) triplets includes irrelevant or redundant information, diminishing performance. Therefore, using \(k = 6\) represents the optimal balance/sweet spot for enriching the model with necessary knowledge while focusing on pertinent information.


\subsection{Effect of Similarity Metric}
\label{effect_retrival}

We further conduct an ablation study to evaluate the effect of different retrieval metrics on explicit commonsense knowledge retrieval across different datasets and input sources using GPT4-O model. We explore the performance of three retrieval metrics: \textit{Cosine Distance}, \textit{Manhattan Distance}, and \textit{Euclidean Distance} to understand their influence on retrieval efficacy across these datasets and input types.

\begin{figure}[!t]
    \centering
    \includegraphics[width = \linewidth]{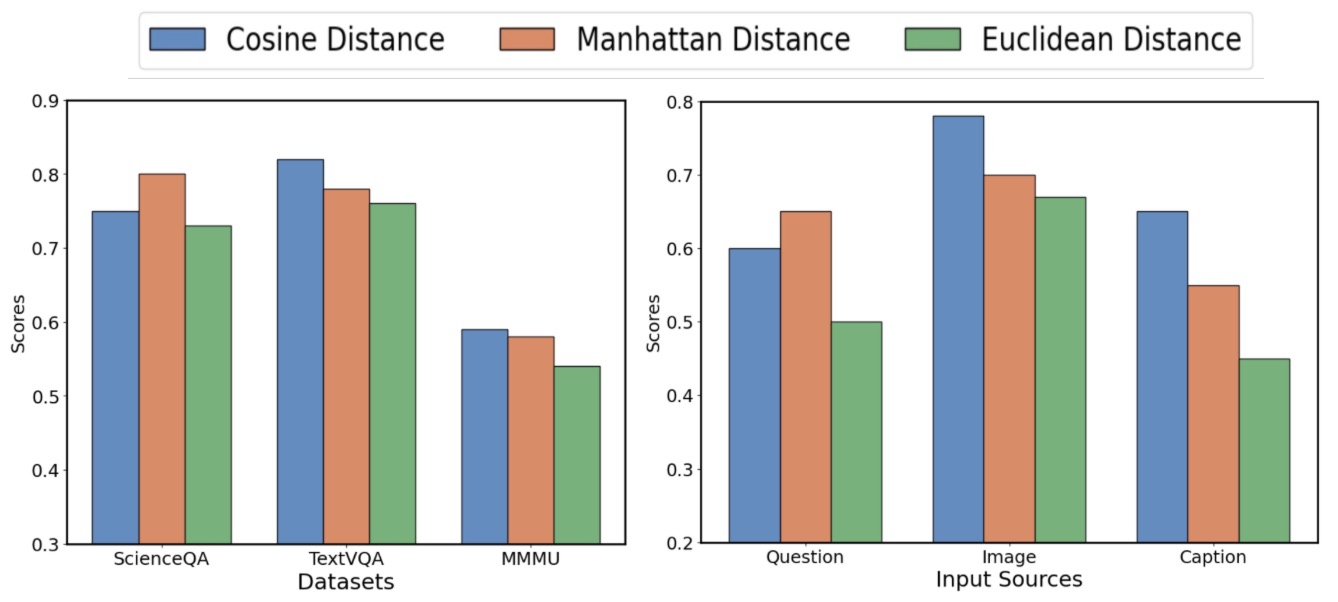}
    \caption{Comparison of accuracy across three datasets (ScienceQA, TextVQA, MMMU) using different distance metrics (Cosine, Manhattan, Euclidean) for explicit commonsense retrieval (Left). The right panel illustrates the performance of these metrics across different input sources (Question, Image, Caption) within the ScienceQA dataset.}
    \label{fig:Retrieval Metric}
\end{figure}

As suggested on the right Figure~\ref{fig:Retrieval Metric}, the performance trends reveal notable differences in retrieval effectiveness depending on the metric and dataset. For ScienceQA, \textit{Manhattan Distance} achieves the highest scores. Meanwhile, the MMMU dataset shows relatively low and uniform scores across all metrics, suggesting that this dataset’s retrieval performance is less sensitive to the choice of metric, potentially due to the diversity and complexity of MMMU’s multimodal inputs.

When comparing retrieval metrics across different input sources, we observe further variations. For question-based retrieval, \textit{Manhattan Distance} consistently yields higher performance scores, indicating that the absolute differences in feature spaces may be more informative for question-centric retrieval. In contrast, image-based and caption-based retrieval achieves the highest scores with \textit{Cosine Distance}, suggesting that angle-based similarity is more effective for capturing visual context in the knowledge graph. 

\subsection{VLPM-Style Fine-Tuning Results}
\label{VLPM_style}

We further perform VLPM-style fine-tuning \cite{long2022gradual, liegoprivacy, wang2025debt} to investigate the effectiveness of implicit multimodal commonsense in MAGIC-VQA framework. Table \ref{tab:VLPM_results} presents a quantitative analysis of the impact of different input nodes—image (I), question (Q), and generated caption text (C)—on the performance of two baseline models (VILT and ALBEF) in VLPM-style fine-tuning. Each node type corresponds to a specific input modality, and when removed, the original embeddings from the pre-trained models are used instead of GCN-trained node embeddings. It is found that including all nodes consistently yields the best outcomes, underscoring the complementary contributions of visual, textual and caption inputs multimodal reasoning. 

Notably, the question node proves to be the most crucial across all nodes. For example, ALBEF’s accuracy on ScienceQA declines from 68.33\% to 57.42\% when the question node is excluded, highlighting its essential role in guiding the model’s attention toward relevant aspects of the image and improving reasoning. On the other hand, the image node also plays a significant role in performance. Visual inputs provide critical scene-level information, enabling models to capture object attributes and spatial relationships, which cannot be fully compensated by text-based inputs alone. 

\begin{table}[!t]
\centering
\renewcommand{\arraystretch}{1.3} 
\setlength{\tabcolsep}{5pt} 
\definecolor{lightgray}{gray}{0.92} 
\begin{adjustbox}{width=\linewidth} 
\begin{tabular}{lccc|ccc}
\hline
\noalign{\hrule height 1.2pt}
\textbf{Model} & \textbf{I} & \textbf{Q} & \textbf{C} & \textbf{SQA} & \textbf{MMMU} & \textbf{TextVQA} \\ 
\hline
\rowcolor{lightgray} ViLT & - & - & - & 56.14 & 23.04 & 41.49 \\ 
MAGIC-VQA\(_{(ViLT)}\) & \xmark & \cmark & \cmark & 60.53 & 20.12 & 40.13 \\ 
MAGIC-VQA\(_{(ViLT)}\) & \cmark & \xmark & \cmark & 53.32 & 14.28 & 32.24 \\ 
MAGIC-VQA\(_{(ViLT)}\) & \cmark & \cmark & \xmark & 63.45 & 22.37 & 43.98 \\ 
\textbf{MAGIC-VQA\(_{(ViLT)}\)} & \textbf{\cmark} & \textbf{\cmark} & \textbf{\cmark} & \textbf{65.41} & \textbf{23.35} & \textbf{44.12} \\ 
\hline
\rowcolor{lightgray} ALBEF & - & - & - & 59.12 & 25.38 & 39.27 \\ 
MAGIC-VQA\(_{(ALBEF)}\) & \xmark & \cmark & \cmark & 61.24 & 24.43 & 37.28 \\ 
MAGIC-VQA\(_{(ALBEF)}\) & \cmark & \xmark & \cmark & 57.42 & 17.21 & 28.42 \\ 
MAGIC-VQA\(_{(ALBEF)}\) & \cmark & \cmark & \xmark & 66.79 & 26.91 & 42.88 \\ 
\textbf{MAGIC-VQA\(_{(ALBEF)}\)} & \textbf{\cmark} & \textbf{\cmark} & \textbf{\cmark} & \textbf{68.33} & \textbf{27.32} & \textbf{43.25} \\ 
\hline
\noalign{\hrule height 1.2pt}
\end{tabular}
\end{adjustbox}
\caption{Combined Results with Multimodal Contributions. The green checkmarks (\cmark) denote the inclusion of a component, while the red crosses (\xmark) denote its exclusion.}
\label{tab:VLPM_results}
\end{table}

\subsection{Additional Results on Knowledge Intensive Benchmarks}
We further attach the result of MAGIC-VQA on two knowledge-intensive tasks including VCR and A-OKVQA as a further complementary analysis.

\begin{figure}[t!]
    \centering
    \includegraphics[width=1\linewidth]{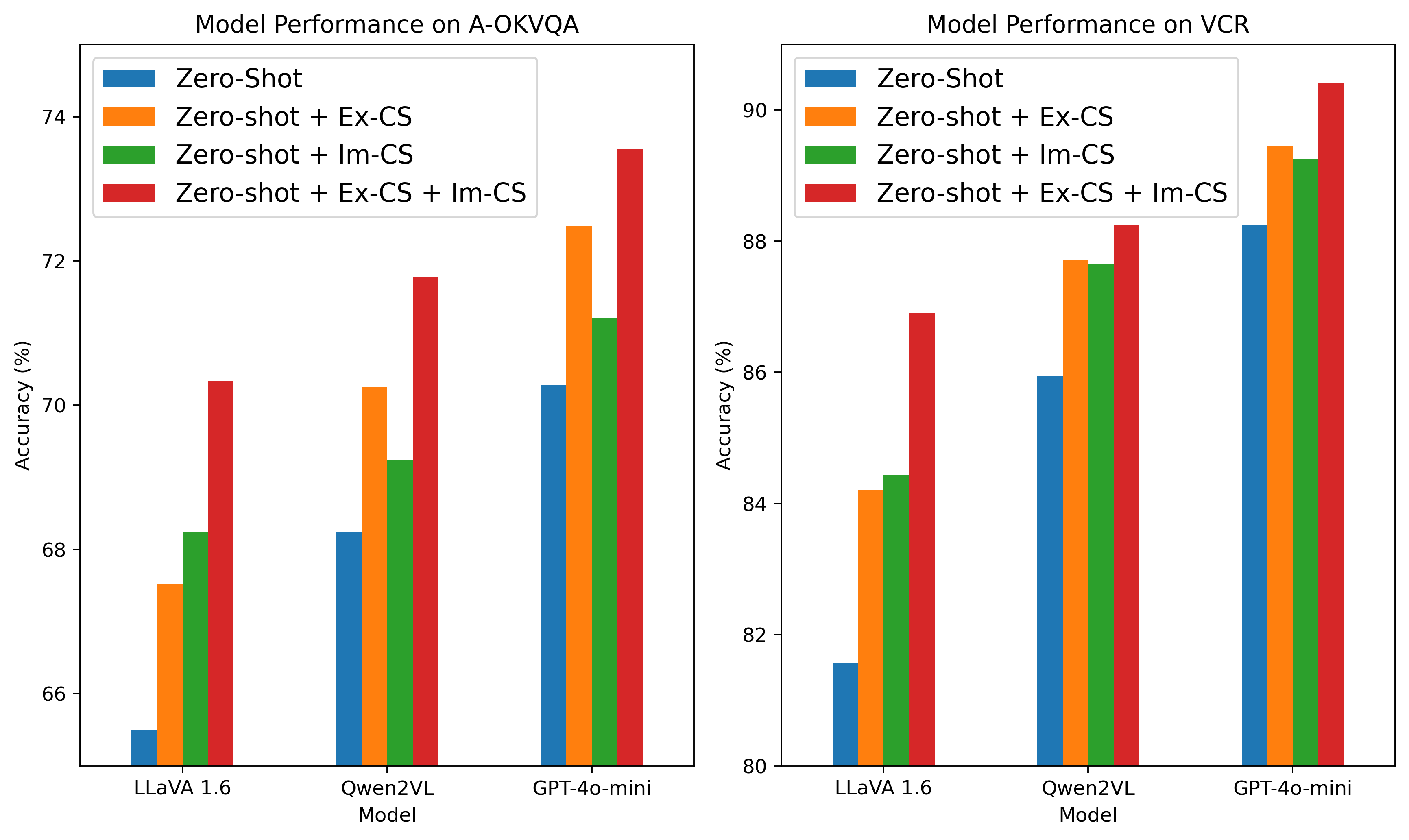}
    \caption{Further Experiment results on VCR and A-OKVQA}
    \label{fig:enter-label}
\end{figure}

\section{Additional Case Studies}
\label{additional case studies}

\begin{figure}[!t]
    \centering
    \begin{subfigure}[b]{\linewidth}
        \centering
        \includegraphics[width=\linewidth]{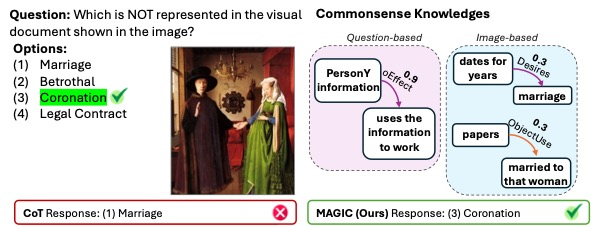}
        \caption{The case from MMMU.}
        \label{fig:subfig1}
    \end{subfigure}
    \vspace{0.5cm} 
    \begin{subfigure}[b]{\linewidth}
        \centering
        \includegraphics[width=\linewidth]{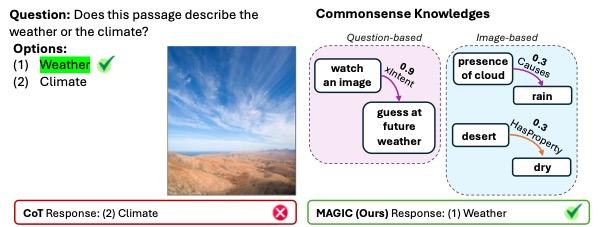}
        \caption{The case from ScienceQA.}
    \end{subfigure}
    \caption{Visualisations of MAGIC-VQA results on the MMMU and ScienceQA datasets, showcasing the role of image-based commonsense knowledge in deriving correct answers. This highlights the cases when image-based commonsense knowledge is more influential in finding the answer.}
    \label{fig:imagekg_essention}
\end{figure}

\begin{figure}[!t]
    \centering
    \begin{subfigure}[b]{\linewidth}
        \centering
        \includegraphics[width=\linewidth]{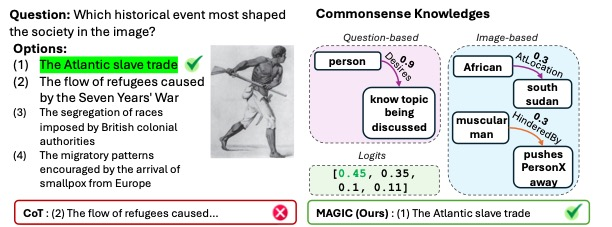}
        \caption{The case from MMMU.}
    \end{subfigure}
    \vspace{0.5cm} 
    \begin{subfigure}[b]{\linewidth}
        \centering
        \includegraphics[width=\linewidth]{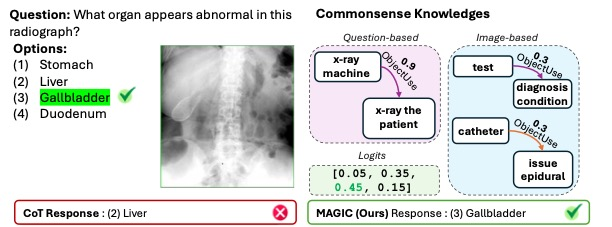}
        \caption{The case from MMMU.}
    \end{subfigure}
    \vspace{-1.0cm}
    \caption{Visualisation of MAGIC-VQA results on MMMU datasets. This highlights the cases when implicit commonsense-based confidence plays an essential role.}
    \label{fig:implicit_kg_essential}
\end{figure}

\begin{figure}[!t]
    \centering
    \begin{subfigure}[b]{\linewidth}
        \centering
        \includegraphics[width=\linewidth]{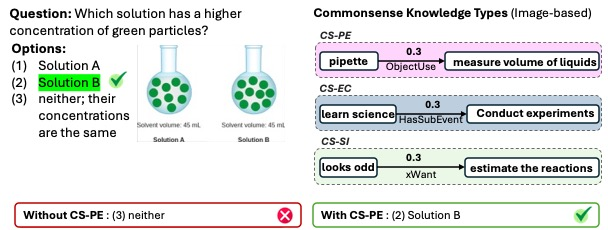}
        \caption{The ScienceQA case when CS-PE is influential}
        \label{fig:case06}
    \end{subfigure}
    \begin{subfigure}[b]{\linewidth}
        \centering
        \includegraphics[width=\linewidth]{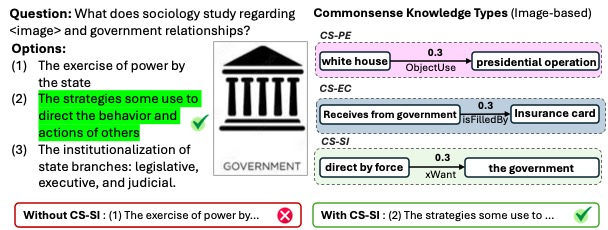}
        \caption{The ScienceQA case when CS-SI is influential}
        \label{fig:case07}
    \end{subfigure}
    \begin{subfigure}[b]{\linewidth}
        \centering
        \includegraphics[width=\linewidth]{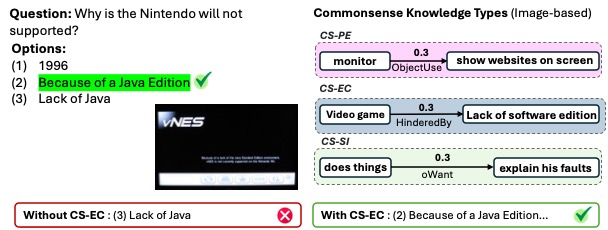}
        \caption{The TextVQA case when CS-PE is influential}
        \label{fig:case09}
    \end{subfigure}
    \caption{Visualisation of MAGIC-VQA results on ScienceQA and TextVQA datasets. This highlights the cases when by-type commonsense knowledge post-processing plays an important role}
    \label{fig:bytype_essential}
\end{figure}

We summarize different qualitative analysis case studies in three types: \textbf{1)} The cases when the image-based explicit commonsense knowledge plays an essential role (Figure \ref{fig:imagekg_essention}), \textbf{2)} The cases when implicit commonsense-based confidence plays an essential role (Figure \ref{fig:implicit_kg_essential}), and \textbf{3)} the cases when by-type commonsense knowledge post-processing plays an important role (Figure \ref{fig:bytype_essential}).

\section{Commonsense Relation Transformation Table}
\label{commonsense-transformation}


We use\textit{ "Someone"},  \textit{"Someone's"} to replace the \textit{"PersonX"},\textit{ "PersonX's"} and \textit{"Another"}, \textit{"Another one's"} to replace \textit{"PersonY"}, \textit{"PersonY's"} separately,  in the heads and tails of the Atomic2020 triplets to enhance clarity and coherence in commonsense-grounded inference.

\begin{table}[t!]
\centering
\small 
\renewcommand{\arraystretch}{1} 
\setlength{\tabcolsep}{4pt} 
\adjustbox{max width=\columnwidth}{ 
\begin{tabular}{@{}p{3cm}p{4.2cm}@{}}
\toprule
\textbf{Relation} & \textbf{Transformed Format} \\
\midrule
\multicolumn{2}{c}{\textit{\textbf{Physical-Entity}}} \\ 
ObjectUse & is used for \\
AtLocation & is at \\
MadeUpOf & is made up of \\
HasProperty & can be \\
CapableOf & is capable of \\
Desires & desires \\
NotDesires & does not desire \\
\midrule
\multicolumn{2}{c}{\textit{\textbf{Event-Centered}}} \\ 
IsAfter & occurs after \\
HasSubEvent & has sub-event \\
IsBefore & occurs before \\
HinderedBy & is hindered by \\
Causes & causes \\
xReason & is because someone \\
isFilledBy & is filled by \\
\midrule
\multicolumn{2}{c}{\textit{\textbf{Social-Interaction}}} \\ 
xNeed & then someone needs \\
xAttr & then someone has attributes \\
xEffect & then someone has the effect \\
xReact & then someone reacts with \\
xWant & then someone wants \\
xIntent & then someone intends \\
oEffect & then the effect on another is \\
oReact & then another reacts with \\
oWant & then another one wants \\
\bottomrule
\end{tabular}
}
\caption{Transformation template of commonsense relations into natural language phrases}
\end{table}


\section{Baselines} \label{Baseline details}

\begin{itemize}
  \item \textbf{LLaVA-1.6 (LLaVA-Next)} \cite{li2024llava}: is an open-source Large Multimodal Model (LMM) designed for enhanced visual and conversational understanding built upon LLaVA \cite{liu2024visual}. It supports higher input resolutions (up to 672x672 pixels) for finer visual detail recognition and incorporates improved visual instruction tuning for better reasoning and OCR capabilities. LLaVA 1.6 is highly efficient, using fewer than 1 million visual instruction tuning samples and a streamlined training process. Its versatility enables it to handle a wide range of applications, from image and text-based tasks to complex multimodal interactions, all while maintaining a minimalist and data-efficient design. We use \href{https://huggingface.co/llava-hf/llava-v1.6-mistral-7b-hf}{llava-v1.6-mistral-7b-hf} checkpoints for zero-shot testing.

  \item \textbf{InternVL2} \cite{chen2024far}: is a state-of-the-art multimodal large model developed by OpenGVLab. It integrates image, video, text, speech, and 3D data, supporting over 100 tasks with exceptional performance across benchmarks. InternVL2 leverages a progressive alignment training strategy and have achieved outstanding results in complex multimodal understanding tasks, rivaling leading commercial closed-source models like GPT-4V \cite{openai2024gpt4technicalreport}. It introduces innovations like vector linking for diverse outputs. It also has parameter sizes ranging from 1B to 76B optimized for efficiency, which delivers high performance even on limited resources. We use \href{https://huggingface.co/OpenGVLab/InternVL2-8B}{InternVL2-8B-hf} checkpoints for zero-shot testing.
  
  \item \textbf{xGen-MM(BLIP-3)} \cite{xue2024xgen}: is a cutting-edge framework for Large Multimodal Models (LMMs) developed by Salesforce AI Research. It features a modular architecture with a scalable vision token sampler and a pre-trained language model, optimized for diverse multimodal tasks such as image captioning, visual question answering, and OCR. It simplifies training objectives with a unified auto-regressive loss and incorporates post-training techniques like Direct Preference Optimization (DPO) and safety fine-tuning to improve truthfulness and mitigate harmful behaviors. We use \href{https://huggingface.co/Salesforce/xgen-mm-phi3-mini-instruct-r-v1}{xgen-mm-phi3-mini-instruct-r-v1} checkpoints for zero-shot testing.
  
  \item \textbf{Qwen2-VL} \cite{wang2024qwen2}: is a cutting-edge vision-language model designed with a robust technical architecture to process multimodal inputs efficiently. It integrates a 675M-parameter Vision Transformer (ViT) enhanced with a Naive Dynamic Resolution mechanism, enabling adaptive encoding of images and videos into variable-length visual tokens to capture detail at multiple scales. To align spatial and temporal information, the model employs Multimodal Rotary Position Embedding (M-RoPE), decomposing positional information into temporal, height, and width dimensions. It also leverages dynamic sequence lengths and efficient parallelism techniques, allowing for deployment in sizes of 2B, 7B, and 72B parameters. We use \href{https://huggingface.co/Qwen/Qwen2-VL-7B-Instruct}{Qwen2-VL-7B-Instruct} checkpoints for zero-shot testing.

  \item \textbf{GPT-4o}~\cite{openai2024gpt4ocard}: is an advanced autoregressive model that processes and generates multimodal content, including text, images, audio, and video, using a unified neural network architecture. It offers significant enhancements in vision and audio understanding, multilingual text generation, and operational efficiency.  The model's training incorporates diverse public and proprietary datasets across modalities, with rigorous post-training alignment to ensure safety and mitigate risks such as bias, misinformation, and unauthorized content generation. We use \href{https://platform.openai.com/docs/guides/vision}{gpt-4o-2024-08-06} checkpoints for zero-shot testing.
\end{itemize}

\section{Implementation Details} \label{implementation details}
We set $K = 30$ to retrieve explicit commonsense knowledge. For by-type commonsense knowledge processing, we configure $\varepsilon = 0.1$ and $k = 6$ to effectively integrate rich commonsense knowledge while minimizing the introduction of excessive noise. For implicit commonsense confidence augmentation,  we follow the default setup in \citet{kipf2017semisupervisedclassificationgraphconvolutional} to explore a standard two-layer GCN. The dimension of the hidden size is set to be 256 and 512, each followed by a dropout layer with the rate to be 0.4. To train the teacher model, we explore batch size to be 64, learning rate to be 1e-5 and epoch to be 30 with early stopping for all models and datasets. There is a global average pooling layer and a output layer using the softmax function after the last GCN layer.

All experiments are conducted on a workstation equipped with one A100 GPU with 40 GB of VRAM. We utilize PyTorch 1.10.0 for model training and the HuggingFace Transformers library for accessing pre-trained models. Our code was written in Python 3.8, and CUDA 11.2 was used for GPU acceleration.

\section{Concrete Example of Input Prompt}
\label{Concrete Example of Input Prompt}

We further include a concrete prompt example of our MAGIC-VQA framework. as demonstrated in  in Table \ref{tab:example-methodology-expanded}, the input sample is augmented with both explicit and implicit commonsense knowledge providing the background information. We  ask the model to first generate the rational then answer the question. 

\begin{table*}[!t]
\centering
\begin{tabular}{p{0.95\linewidth}}
\toprule
\textbf{Background} \\ \midrule
You are an advanced Vision-Language Model assistant designed to answer multiple-choice questions based on a given image. Your task is to select the most appropriate option from the provided answer choices. You are given an input image, a question related to the image, the image caption, multiple-choice answer options, and both explicit and implicit commonsense knowledge. \\
\\

Explicit commonsense knowledge consists of statements related to the input, categorized as image-related commonsense, question-related commonsense, and caption-related commonsense. Implicit commonsense knowledge includes the relevance level (e.g., highly relevant, relevant, less relevant) assigned to each explicit commonsense statement and the confidence of each candidate option, where higher values indicate a greater likelihood of being correct. \\
\\
Your objective is to integrate the explicit and implicit commonsense knowledge with the provided information to generate a step-by-step reasoning. Based on this rationale, you will select the most appropriate answer from the given options. \\ \midrule

\textbf{Input Information} \\ \midrule
\textbf{Image:} \\ 
\includegraphics[width=0.5\linewidth]{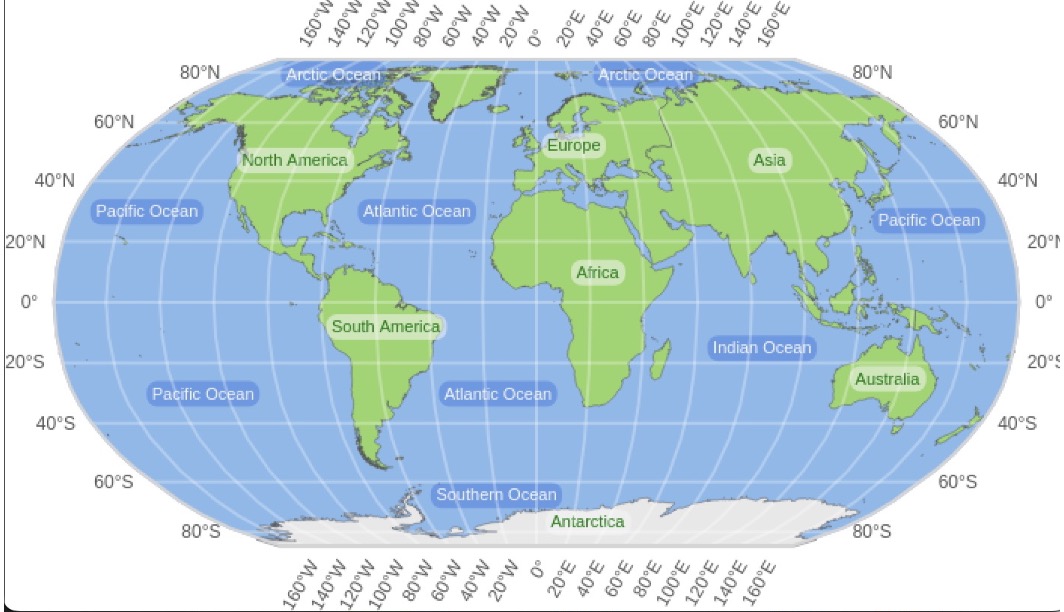} \\

\textbf{Question:} Which of these oceans does the prime meridian intersect? \\
\textbf{Caption:} An image of a world map with labeled continents and oceans. \\
\textbf{Options:} \\
\hspace{1em} A. "the Atlantic Ocean" \\
\hspace{1em} B. "the Indian Ocean" \\
\hspace{1em} C. "the Pacific Ocean" \\ \midrule

\textbf{Explicit Commonsense Knowledge} \\ \midrule
\textit{Image-Related Commonsense:} \\
\hspace{1em} - The Atlantic Ocean is at the western hemisphere. (Highly Relevant) \\
\hspace{1em} - A world traveler is capable of crossing many time zones. (Relevant) \\

\textit{Question-Related Commonsense:} \\
\hspace{1em} - A traveler is capable of crossing geographical borders. (Highly Relevant) \\
\hspace{1em} - Someone who is far from home might want to measure the distance. (Less Relevant) \\

\textit{Caption-Related Commonsense:} \\
\hspace{1em} - The Atlantic Ocean is used for separating continents. (Highly Relevant) \\
\hspace{1em} - If someone sees the ocean, they might think of traveling to it. (Relevant) \\ \midrule

\textbf{Implicit Commonsense Knowledge (Confidence for Each Option)} \\
\hspace{1em} A: 0.6 \\
\hspace{1em} B: 0.05 \\
\hspace{1em} C: 0.35 \\ \midrule

\textbf{Rationale:} 
 \\ \midrule

\textbf{Answer:} 
\\ \bottomrule
\end{tabular}
\vspace{-0.3cm}
\caption{A concrete example of the input prompt}
\label{tab:example-methodology-expanded}
\end{table*}

\section{Commonsense Category Analysis Prompt Format} 
\label{Commonsense Category Analysis Prompt Format}

To analyze the commonsense knowledge distribution  within each selected dataset, we provide the following prompt template to classify each sample to their most relevant commonsense knowledge (CS-PE, CS-EC, CS-SI) covered in Table~\ref{tab:commonsense_analysis_prompt_template}.

\begin{table}[]
\centering
\begin{tabular}{|p{0.9\linewidth}|} 
\hline
\textbf{Prompt Template for Commonsense Category Classification} \\
\hline
\begin{minipage}{\linewidth} 
\textbf{Instructions:}

You are an expert in commonsense reasoning and knowledge representation. Your task is to classify each sample into one of three commonsense categories:

1. \textbf{Physical-Entity Commonsense (CS-PE)}: Knowledge about physical objects, their properties, uses, locations, and physical attributes. This includes understanding what things are made of, typical or atypical uses, and physical characteristics. 

2. \textbf{Event-Centered Commonsense (CS-EC)}: Knowledge about events, including their causes, effects, prerequisites, sequences, and hindrances. This encompasses understanding how events are related in time and causality. 

3. \textbf{Social-Interaction Commonsense (CS-SI)}: Knowledge about social behaviors, mental states, interactions, and interpersonal dynamics. This involves understanding intentions, emotional reactions, and attributes in social contexts.

\textbf{Sample:}

\begin{itemize}
    \item \textbf{Image:} \( <Image Caption>\)
    \item \textbf{Question:} \( <Question>\)
    \item \textbf{Choices:} \( <Options>\)
    \item \textbf{Answer:} \( <Answer>\)
\end{itemize}

\textbf{Reasoning Steps:}

Please first examine the question and answer choices, along with the image caption, to identify the main focus of the sample. Then provide a step-by-step reasoning on how specific elements of the sample align with the potential commonsense category. Then assign the appropriate commonsense category (CS-PE, CS-EC, or CS-SI) based on the provided rationale.

\textbf{Classification:}
\\

\end{minipage} \\
\hline
\end{tabular}
\caption{Prompt Template for Classifying Samples into Commonsense Categories}
\label{tab:commonsense_analysis_prompt_template}
\end{table}

\section{Retrieved Commonsense Triplet Cosine Similarity Distribution}
\begin{figure*}[!b]
    \centering
    \begin{subfigure}[b]{\textwidth} 
        \centering
        \includegraphics[width=0.95\textwidth, height=0.22\textheight]{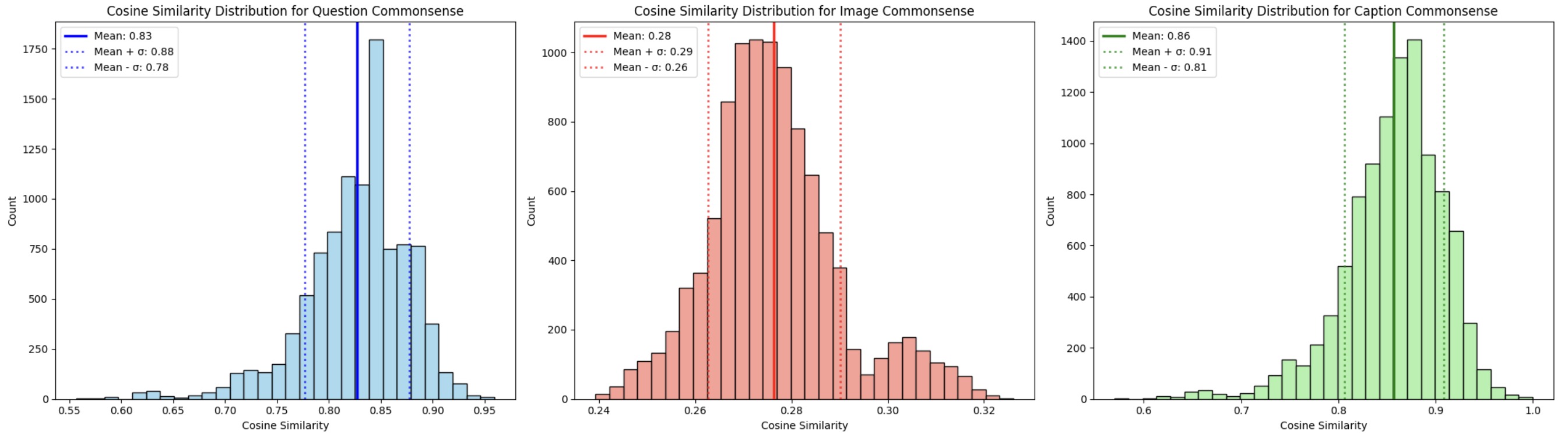}
        \caption{ScienceQA}
        \label{fig:scienceqa}
    \end{subfigure}
    \begin{subfigure}[b]{\textwidth}
        \centering
        \includegraphics[width=0.95\textwidth, height=0.22\textheight]{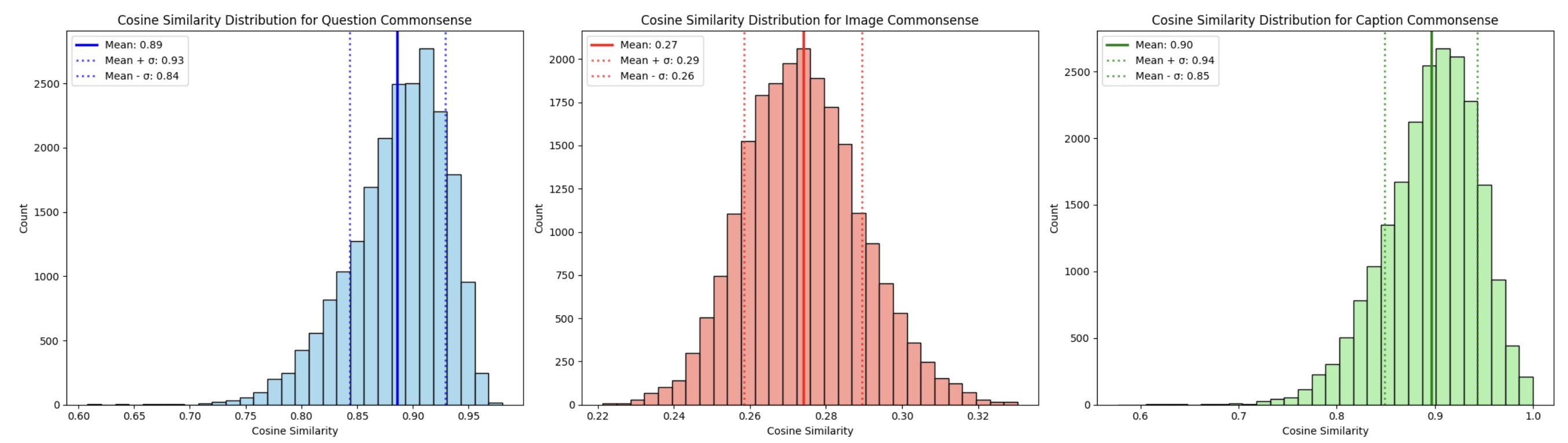}
        \caption{TextVQA}
        \label{fig:textvqa}
    \end{subfigure}
    \begin{subfigure}[b]{\textwidth}
        \centering
        \includegraphics[width=0.95\textwidth, height=0.22\textheight]{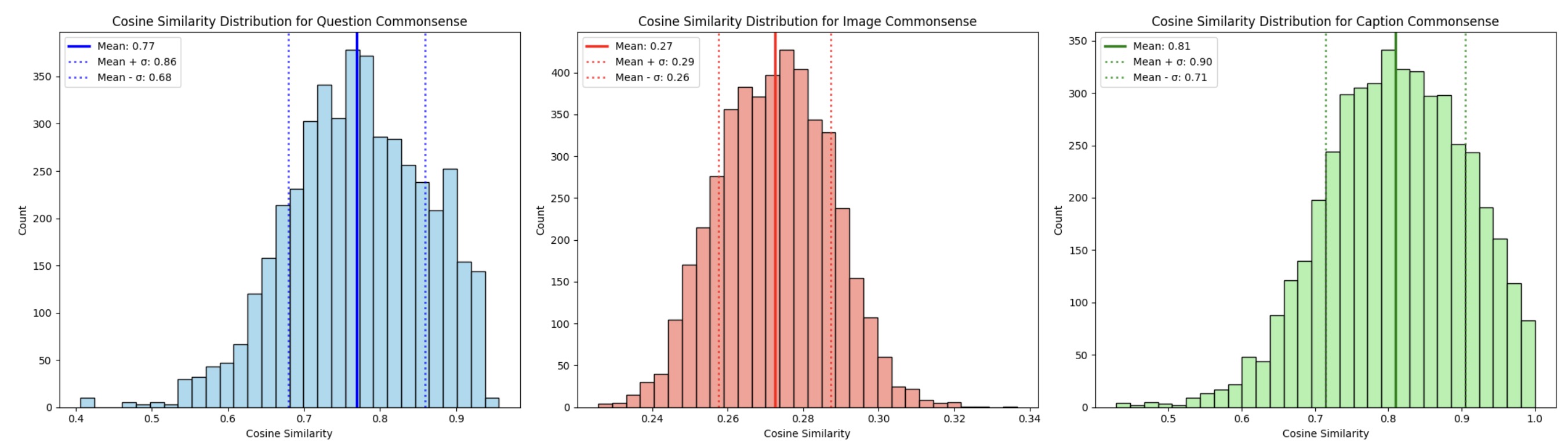}
        \caption{MMMU}
        \label{fig:mmmu}
    \end{subfigure}
    \caption{Overall cosine similarity distributions for three input sources within each dataset. The first column represents the cosine similarity distribution of retrieved triplets for input question. The second column represents the cosine similarity distribution of retrieved triplets for input image. The third column represents the cosine similarity distribution of retrieved triplets for input caption.}
    \label{fig:Overall caption describing all three distributions}
\end{figure*}

Figure \ref{fig:Overall caption describing all three distributions} depicts the cosine similarity distribution of the retrieved triplets of each input source across all three datasets.

\end{document}